\documentclass{article}

\PassOptionsToPackage{numbers, sort, comma, square}{natbib}

\usepackage[main, final]{neurips_2026}

\usepackage{authblk}
\usepackage[utf8]{inputenc} 
\usepackage[T1]{fontenc}    
\usepackage{url}            
\usepackage{booktabs}       
\usepackage{amsfonts}       
\usepackage{nicefrac}       
\usepackage{microtype}      
\usepackage[table]{xcolor}  

\usepackage{amsmath}
\usepackage{amsfonts}
\usepackage{amssymb}
\usepackage{graphicx}
\usepackage{subfigure}
\usepackage{booktabs}
\usepackage{multirow}
\usepackage{amsthm}
\usepackage[ruled]{algorithm2e}
\usepackage{upgreek}
\usepackage{appendix}
\usepackage{placeins}
\usepackage{hyperref}
\hypersetup{
	hidelinks,
	pdfstartview=Fit,
	breaklinks=true}

\DeclareRobustCommand{\secref}[1]{\hyperref[#1]{Section~\ref*{#1}}}
\DeclareRobustCommand{\appref}[1]{\hyperref[#1]{Appendix~\ref*{#1}}}
\DeclareRobustCommand{\figref}[1]{\hyperref[#1]{Figure~\ref*{#1}}}
\DeclareRobustCommand{\figsrefrange}[2]{\hyperref[#1]{Figures~\ref*{#1}}\hyperref[#2]{--\ref*{#2}}}
\DeclareRobustCommand{\figrefshort}[1]{\hyperref[#1]{Fig.~\ref*{#1}}}
\DeclareRobustCommand{\tabref}[1]{\hyperref[#1]{Table~\ref*{#1}}}
\DeclareRobustCommand{\eqreftext}[1]{\hyperref[#1]{Eq.~\ref*{#1}}}
\DeclareRobustCommand{\eqsrefrange}[2]{\hyperref[#1]{Eqs.~\ref*{#1}}\hyperref[#2]{--\ref*{#2}}}

\title{Is Class Signal Clustered or Routed in Task-Induced Implicit Neural Representation Weight Spaces?}

\author[${\ast}$,1]{\textbf{Xinyi Guo}\thanks{Contributed equally}}
\author[$\boldsymbol{\ast}$,2]{\textbf{Mingyi He}}
\author[$\boldsymbol{\ast}$,1]{\textbf{Haobin Ding}}
\author[3]{\textbf{Weiming Chen}}
\author[3]{\textbf{Xinrui Chen}}
\author[3]{\textbf{Jiawen Li}}
\author[4]{\textbf{Di Zhang}}
\author[3]{\textbf{Minxi Ouyang}}
\author[3]{\textbf{Yizhi Wang}}
\author[$\dagger$,3]{\textbf{Xitong Ling}\thanks{Corresponding authors}}

\affil[1]{%
  \textup{South China Normal University}}
\affil[2]{%
  \textup{Beijing University of Chemical Technology}}
\affil[3]{%
  \textup{Tsinghua University}}
\affil[4]{%
  \textup{Xi'an Jiaotong University}}

\date{} 
\begin{document}

	\maketitle

	\begin{abstract}
		Implicit neural representations (INRs) encode images as neural-network weights, making image classification a problem of weight-space classifiability.
        A natural geometric hypothesis is that classifier feedback should make image-specific weights cluster by class in the shared-anchor coordinate.
        We test this hypothesis in the SIREN-based Meta Weight Transformer (MWT) regime, where end-to-end training meta-learns a shared initialization and inner-loop update schedule for fitting image-specific SIRENs.
        We find that this prediction fails.
		Exposed weight-space geometry and supervised clustering pressure do not reliably track trained-reader accuracy; clustering can even make local neighborhoods more class-consistent while making the trained reader worse.
		Crucially, the reader constructs rather than inherits class-aligned geometry: token-flow diagnostics show that class-aligned neighborhoods become strongly predictive of trained-reader accuracy only after late reader interactions, not in the input coordinate.
        We further identify the native SIREN bias column in the augmented weight token as a low-dimensional, sample-dependent causal readout route for the trained reader; targeted controls rule out generic scalar-column and marginal-distribution artifacts.
        The diagnosis motivates interventions that strengthen reader routing, add an explicit bias route, or use denser inner-loop fitting; under the lane-specific training conventions used here, route-directed variants often outperform the shared-anchor baseline but interact non-additively.
        Task-induced INR weights are classifiable not because they form raw geometric clusters, but because their class signal is routed through the reader.
	\end{abstract}

	\section{Introduction}
    Neural-network weights are increasingly treated as a data modality rather than merely as optimization endpoints.
    Model-zoo and hyper-representation work showed that trained weights can be characterized and embedded directly \citep{eilertsen2020classifying,unterthiner2020predicting,schurholt2021hyperrepresentations,schurholt2022modelzoos}; neural-functional and graph-based readers then made structure and symmetry explicit modeling concerns \citep{navon2023dwsnet,zhou2023nfn,zhou2023nft,kofinas2024neuralgraphs,lim2024gmn}.
    The weight-space learning view consolidates understanding, representation, and generation directions \citep{schurholt2022hypergenerative,peebles2022gpt,erkoc2023hyperdiffusion,han2026wsl}, but also exposes a tension: useful task structure in weights need not be raw coordinate geometry.

    Implicit neural representations (INRs) make this tension concrete.
    Images can be encoded as parameters of neural functions; sinusoidal representation networks (SIRENs) provide a standard sinusoidal global MLP for such signals \citep{sitzmann2020siren}.
    Data-to-Functa treats fitted functions as learning objects \citep{dupont2022data}, and INR-zoo benchmarks show that implicit-function weights can form practical datasets \citep{ma2024implicitzoo}.
    This motivates the setting we study: task-induced global SIREN weight spaces, where the question is not merely whether image-specific weights can be classified, but what makes that classifiability usable.

    \begin{figure*}[!t]
    \centering
    \includegraphics[width=0.96\textwidth]{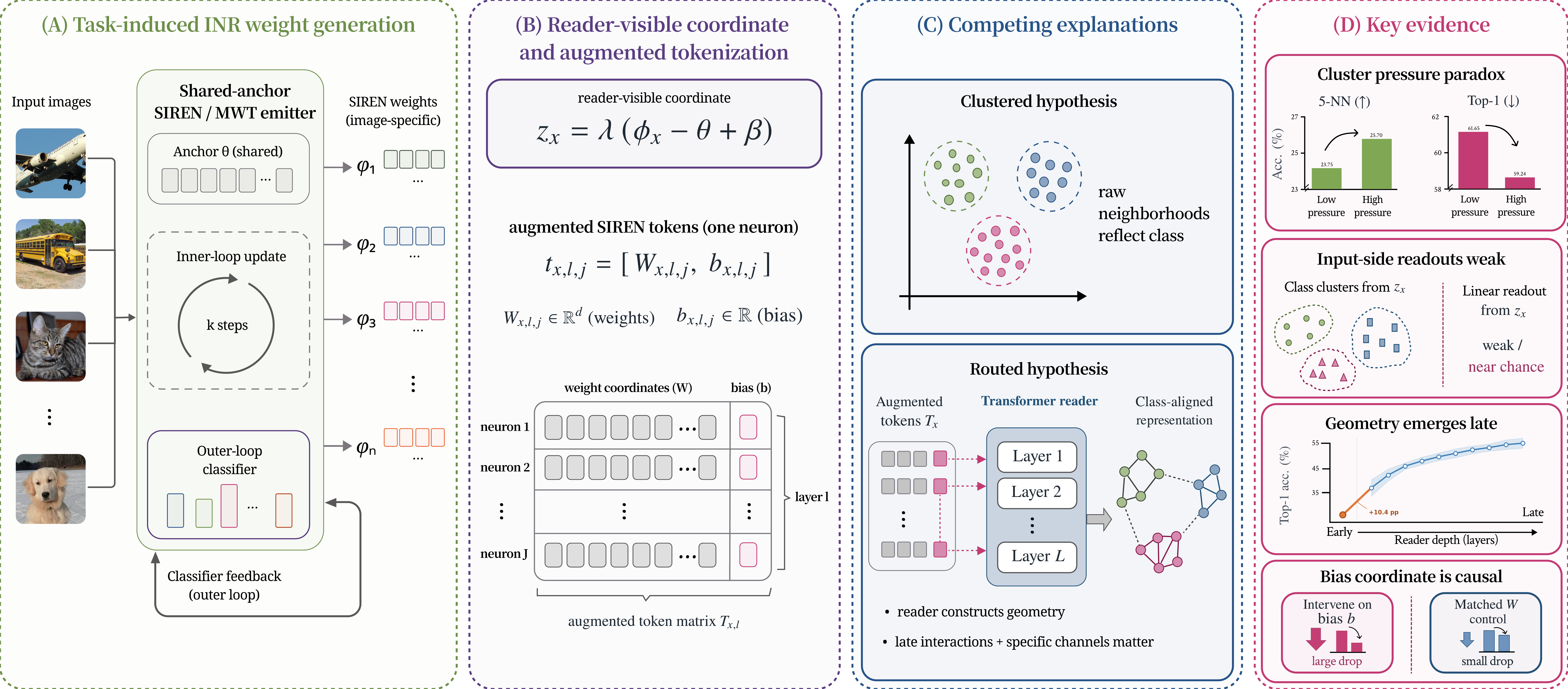}
    \vspace{-1mm}
    \caption{
    \textbf{Overview: diagnosing whether class signal is clustered or routed in task-induced INR weights.}
    The pipeline fits image-specific SIRENs from a shared anchor under classifier feedback, exposes reader-visible weight tokens, and then tests two competing explanations of classifiability.
    The paper contrasts two observables: exposed coordinate geometry, which can be tested directly, and reader-side states, which are later traced to test whether class structure becomes usable inside the trained reader.
    }\label{fig:overview}
    \end{figure*}

    The Meta Weight Transformer (MWT) supplies a stringent testbed.
    Unlike reconstruction-only INR pipelines, it meta-learns a shared SIREN initialization and inner-loop update schedule while classifier feedback shapes the produced image-specific weights \citep{gielisse2025mwt,papa2024fitanef,finn2017maml,li2017metasgd}.
    Because all examples arise from one anchor through a few updates, the native residual coordinate is where raw class geometry should most plausibly appear.
    Residual-update, adapter, and alignment work likewise show that reference coordinates matter for weight comparisons \citep{houlsby2019adapters,ilharco2023taskarithmetic,hu2022lora,entezari2022permutation,ainsworth2023gitrebasin,navon2024deepalign}.
    This makes the shared-anchor coordinate the most favorable place to look for raw semantic geometry.

    The clustered account is therefore a serious hypothesis rather than a strawman: if task feedback writes semantic information directly into the anchored coordinate, raw neighborhoods, component summaries, spectral structure, shallow probes, and supervised clustering pressure should move with trained-reader accuracy.
    Such expectations are not arbitrary; Center Loss and supervised contrastive learning explicitly compact same-class representations, while neural-collapse analyses show that simple class geometry can emerge in supervised representations \citep{wen2016centerloss,khosla2020supcon,papyan2020neuralcollapse}.
    In the task-induced SIREN coordinate, however, this prediction chain breaks.
    The exposed coordinate can satisfy some neighborhood-level tests while still failing to explain the trained reader's task behavior.
    The issue is therefore not whether one can impose class-shaped structure on the coordinate, but whether that structure is the route by which the reader classifies.

    This failure leaves a fork: either classifier feedback fails to induce useful class information in the SIREN weights, or the information exists but is not arranged as directly readable metric geometry.
    We resolve this fork by tracing the trained reader, shifting the explanation from inherited coordinate geometry to routed readout.

    The route is also channel-specific: within MWT's augmented SIREN token, the reader-visible bias coordinate is a native weight-space input \citep{gielisse2025mwt}.
    We use targeted interventions and controls to test whether this coordinate is a genuine readout route rather than a generic scalar-column or distributional artifact.
    Finally, mechanism-guided validation tests whether diagnosed routes are actionable and modular.
    \textbf{Contributions.} We make three contributions: (i) a clustered-versus-routed formulation for task-induced INR weights; (ii) a geometry--classifiability gap that separates exposed coordinate structure from trained-reader behavior; and (iii) a reader-routing account that localizes a reader-visible bias route and tests its non-additive intervention consequences.

	\section{Formulating classifiability in task-induced SIREN weight spaces}\label{sec:problem-formulation}

    Implicit neural representations encode each image as the parameters of a neural function. Here the object of study is not an independently fitted SIREN zoo, but a task-induced SIREN weight space produced by end-to-end MWT training. For an image \(x\), MWT initializes a SIREN at a meta-learned shared anchor \(\theta\) and applies \(k\) reconstruction-driven inner-loop steps,
    \begin{equation}
    \phi_x = U_\alpha^k(\theta; x),
    \qquad
    \Delta_x = \phi_x-\theta ,
    \label{eq:weights-and-residual}
    \end{equation}
    where \(\alpha\) is the learned inner-loop update schedule and \(\Delta_x\) is the anchored residual coordinate. The SIREN \(f_{\phi_x}\) maps image coordinates to RGB values. Unlike two-stage INR classification, MWT back-propagates the classification loss through the fitting procedure, jointly updating the shared initialization, update schedule, and reader. Thus the distribution of \(\phi_x\) is shaped by downstream classification feedback rather than by reconstruction alone \citep{gielisse2025mwt,finn2017maml,li2017metasgd}.

    Because all examples are produced from a shared anchor, \(\Delta_x\) is the natural coordinate for comparing them. For fixed \(\theta\), residual differences preserve pairwise parameter differences, \(\Delta_i-\Delta_j=\phi_i-\phi_j\). We therefore treat \(\Delta_x\) as an anchored coordinate chart of the same SIREN parameter space. This reference-based view is consistent with weight-space manipulation and alignment work, where task vectors, adapters, low-rank updates, and reference-aligned comparisons are defined relative to shared or matched parameterizations \citep{houlsby2019adapters,ilharco2023taskarithmetic,hu2022lora,entezari2022permutation,ainsworth2023gitrebasin,navon2024deepalign}.

    As in MWT, the reader consumes a scaled, shifted anchored coordinate,
    \begin{equation}
    z_x=\lambda(\Delta_x+\beta)
    =\lambda(\phi_x-\theta+\beta).
    \label{eq:mwt-coordinate}
    \end{equation}
    Here \(\beta\in\mathbb{R}^{|\theta|}\) is MWT's learned reader-side per-weight positional shift, and \(\lambda\) is the scalar input scale that brings small residual values into the reader's numerical range. We use \(z_x\) for the reader-visible coordinate throughout the paper; \secref{sec:controls} tests alternative coordinate packagings as controls.

    MWT tokenizes hidden SIREN layers by representing each output neuron as its incoming weights together with its bias. For hidden layer \(\ell\) and output neuron \(j\), we write the reader-visible augmented token schematically as
    \begin{equation}
    t_{x,\ell,j}=[W_{x,\ell,j},\, b_{x,\ell,j}].
    \label{eq:augmented-token}
    \end{equation}
    The last coordinate \(b_{x,\ell,j}\) indexes the SIREN neuron bias parameter inside the augmented token after the same residual/shift/scale packaging as \eqreftext{eq:mwt-coordinate}. It is part of the reader-visible weight data, not a reader parameter. This \(b\)-coordinate is distinct from \(\beta\): \(\beta\) is a learned reader-side positional shift, whereas \(b\) is a SIREN-side bias coordinate within the token. Later controls act on \(b\) while keeping \(\beta\) fixed, so this distinction is essential for interpretation \citep{gielisse2025mwt}.

    Given this coordinate system and tokenization, we compare two observables: trained-reader classifiability and exposed weight-space geometry. Trained-reader classifiability is the Top-1 or Top-5 accuracy of the trained reader \(R_\psi(z_x)\). Exposed geometry diagnostics ask whether class structure is visible before reader-side computation. These include nearest-neighbor class consistency, class-conditional distances, component summaries, spectral statistics, and low-dimensional probes. The two observables answer different questions: accuracy measures end-task classification, whereas the diagnostics measure whether class structure is directly exposed in the native coordinate before reader-side token interactions.

    For the reader-side diagnostics used later, let
    \begin{equation}
    h_x^{(m)} = R_\psi^{(m)}(z_x)
    \label{eq:reader-state}
    \end{equation}
    denote the representation after the \(m\)-th reader block.
    The hypotheses below differ in whether class geometry should already be exposed in \(z_x\), or become readable only in such reader-side states.

    This setup supports two competing accounts of weight-space classifiability.

    \paragraph{Clustered hypothesis.}
    Classifiability is inherited from exposed weight-space geometry. If classifier feedback writes semantic structure directly into the anchored SIREN coordinate, then same-class examples should become locally organized in \(z_x\). This predicts that nearest-neighbor consistency, component summaries, spectral diagnostics, and shallow probes should co-vary with trained-reader accuracy. It also predicts that supervised compactness pressure should help the reader by tightening same-class neighborhoods and increasing class separability \citep{wen2016centerloss,khosla2020supcon,papyan2020neuralcollapse}.

    \paragraph{Routed hypothesis.}
    Classifiability is constructed by reader-side token computation.
    Class-relevant information may be present in \(z_x\) without being arranged as simple metric clusters.
    This predicts that raw geometry diagnostics can fail to track trained-reader accuracy, supervised clustering pressure can improve exposed compactness without improving the trained reader, class-aligned geometry may become readable only in \(h_x^{(m)}\), specific token channels can be causally used by the trained reader, and interventions targeting the same routed signal need not compose additively.

    The tests proceed from exposed geometry to reader-side mechanism, intervention, and scope. \secref{sec:geometry-gap} evaluates whether exposed coordinate geometry explains classifiability. \secref{sec:routing-bias} traces where class-aligned structure becomes readable inside the reader and uses targeted interventions to test whether the trained reader relies on the reader-visible SIREN bias coordinate. \secref{sec:interventions} uses the diagnosis to order interventions and asks whether their gains behave modularly. \secref{sec:controls} bounds the claim with coordinate-packaging and scope controls.

	\section{The geometry--classifiability gap}\label{sec:geometry-gap}

    The clustered account yields one direct test: before reader-side token mixing, the reader-visible coordinate should already exhibit class-consistent local neighborhoods.
    The setting gives this prediction a favorable raw-geometry test: all image-specific SIRENs are produced from a shared MWT anchor, and classifier feedback shapes the emitted weight distribution during training \citep{gielisse2025mwt}.

    \paragraph{Raw-coordinate diagnostics do not explain accuracy.}
    The first test inspects the exposed coordinate before reader-side token interactions.
    The diagnostic suite spans local neighborhoods, class-conditional distances, component, spectral, intrinsic/global, and Gromov/geodesic-style summaries over the offset cloud.
    Across the completed diagnostics, no raw-coordinate metric family monotonically explains validation accuracy; \appref{app:raw-diagnostic-inventory} reports the full diagnostic inventory and per-metric outcomes.

\begin{figure*}[t]
    \centering
    \begin{minipage}[t]{0.31\textwidth}
    \centering
    \includegraphics[width=\linewidth]{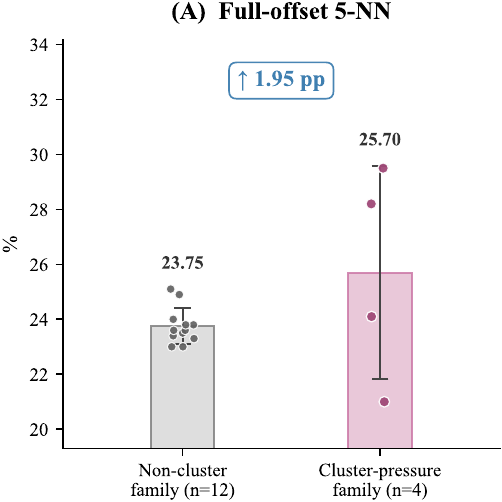}
    \end{minipage}
    \hfill
    \begin{minipage}[t]{0.31\textwidth}
    \centering
    \includegraphics[width=\linewidth]{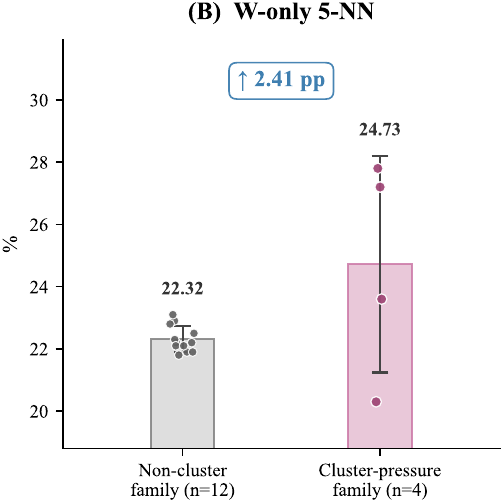}
    \end{minipage}
    \hfill
    \begin{minipage}[t]{0.31\textwidth}
    \centering
    \includegraphics[width=\linewidth]{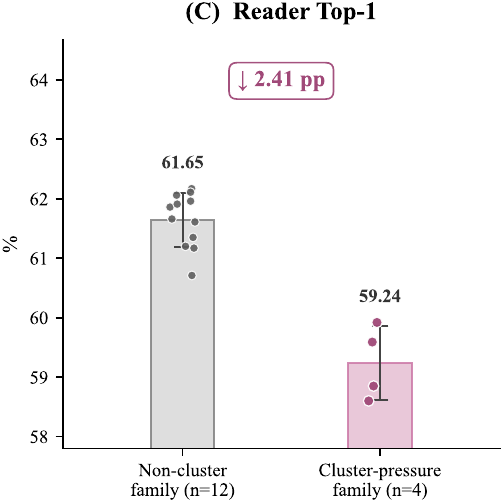}
    \end{minipage}
    \vspace{-1mm}
    \caption{
    \textbf{In the broader family panel, cluster pressure can improve exposed local consistency while lowering trained-reader accuracy.}
    Bars show family means with $\pm1$ sample s.d. across emitter configurations ($n=12$ non-cluster, $n=4$ cluster-pressure); $\Delta$ is cluster-pressure minus non-cluster.
    Full-offset 5-NN rises $23.75{\to}25.70$ and W-only 5-NN rises $22.32{\to}24.73$, while trained-reader Top-1 falls $61.65{\to}59.24$.
    This is a family-level stress test; per-emitter heterogeneity and seed-stability checks are in \appref{app:geometry-heterogeneity}.
    }\label{fig:geometry-gap}
    \end{figure*}

    The gap is therefore one of explanation, not signal absence: the coordinate may carry class-related variation, but exposed raw-coordinate diagnostics do not consistently track the reader's task metric.
    \figref{fig:geometry-gap} turns this distinction into a paired stress test: the same family comparison asks whether improving exposed local consistency also improves the trained reader.

    \paragraph{Local consistency is not reader classifiability.}
    \figref{fig:geometry-gap} shows the paired reversal predicted to be impossible under the simplest clustered account.
    Because the account is a local-neighborhood claim, the first two panels use 5-NN class consistency as the direct exposed-geometry diagnostic; the third panel reads out the trained weight reader.
    Cluster pressure improves full-offset 5-NN by $+1.95$pp and W-only 5-NN by $+2.41$pp, yet the trained reader moves in the opposite direction, with Top-1 decreasing by $2.41$pp.

    Thus \figref{fig:geometry-gap} is a family-level diagnostic reversal rather than a per-emitter guarantee. On the matched six-emitter cross-seed pivot, the raw-side sign is near-neutral rather than reversed ($-0.35$pp at raw-residual PCA-128 5-NN), while the late-reader rows keep the routed-account sign. The Top-1 drop is best read not as evidence against same-class compactness in general, but as misaligned pressure on an exposed pre-mixing statistic, which can trade off with the coordinate patterns later read out by the trained reader. \appref{app:geometry-heterogeneity} reports heterogeneity and seed-stability checks.

\begin{figure*}[t]
    \centering
    \includegraphics[width=0.96\textwidth]{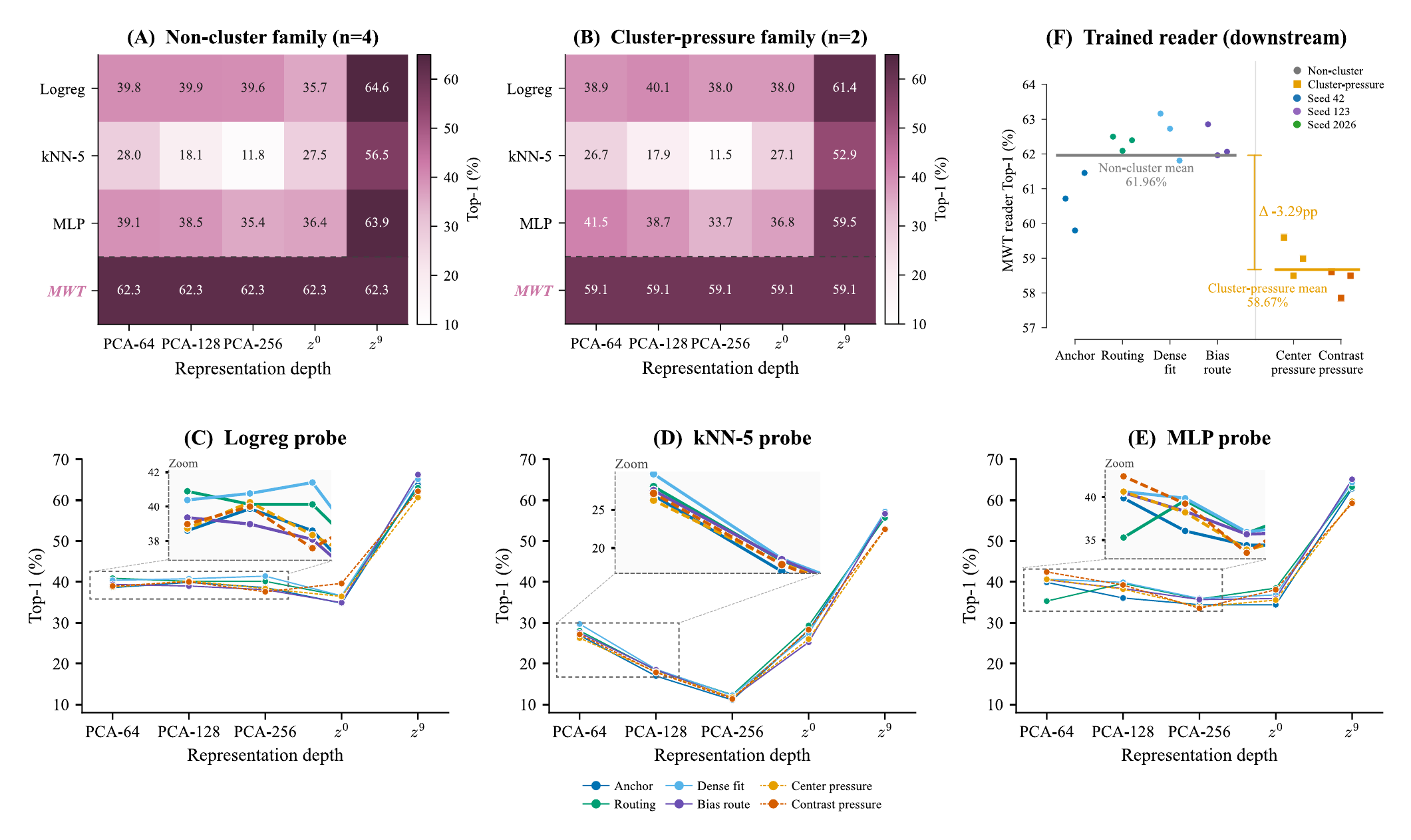}
    \caption{
    \textbf{Input-side readouts do not rescue the clustered account.}
    Top row: shallow input-side probes and the downstream trained weight reader; the MWT row is the trained reader, not an input-side probe.
    Bottom row: per-emitter probe trajectories across representation depth.
    The panel visualizes the input-readout pattern; the prose and \appref{app:input-readouts-full} report the cross-seed pivot used for quantitative contrasts.
    }\label{fig:input-side-no-rescue}
    \end{figure*}

    \paragraph{Input-side readouts do not rescue the clustered account.}
    A weaker clustered account might argue that raw 5-NN is too strict, while a shallow input-side readout could still recover the useful class signal.
    \figref{fig:input-side-no-rescue} tests this weaker account with shallow input-side readouts before comparing them to the trained weight reader.
    For the high-dimensional raw offsets, we use a fixed PCA-128 compression before shallow probes; the purpose is not to optimize a reader, but to ask whether a simple input-side readout already exposes the signal.
    Under cluster pressure, the cross-seed pivot shows mixed and small input-side gaps: raw PCA-128 logreg and 5-NN are $-0.68$pp and $+0.18$pp, while reader-input logreg and 5-NN are $+0.79$pp and $-1.34$pp. Late-reader probes move in the trained-reader direction, with $h^{(9)}$ logreg and 5-NN gaps of $-3.78$pp and $-4.69$pp; the co-trained reader gap is $-3.29$pp on the matched cross-seed pivot.

    The figure therefore rules out a simple probe-capacity rescue: input-side separability is not absent, but it is not the classifiability achieved by the co-trained reader.

    \paragraph{The raw coordinate does not simply expose reconstruction quality.}
    A second alternative is that the input-side coordinate fails for semantics but directly exposes low-level reconstruction quality.
    This is also not supported.
    At the reader-input projection, PSNR ridge prediction remains weak, with $R^2=0.02$--$0.10$ across the six-emitter readout panel.
    Thus the native coordinate does not expose a directly readable reconstruction-quality geometry that merely lacks semantics.
    Both class and reconstruction readouts point to the same conclusion: exposed input-side geometry is not sufficient to explain trained-reader classifiability.

    \paragraph{What the gap establishes.}
    These results establish the gap, not yet the route.
    They show what classifiability is not: it is not an intrinsic property of raw metric neighborhoods, standard coordinate summaries, supervised compactness objectives, or shallow input-side readouts in the exposed coordinate.
    The next section traces where class-relevant structure becomes usable and which reader-visible token channel supports the readout.

    \section{Reader construction and bias-column routing}\label{sec:routing-bias}

    The geometry--classifiability gap shows that the exposed SIREN coordinate is not already organized as the class geometry used by the trained reader.
    It does not imply that the induced SIRENs contain no class-relevant information.
    We now ask where this information becomes usable, and through which reader-visible coordinate channel it is routed.

\begin{figure*}[t]
    \centering
    \includegraphics[width=0.96\textwidth]{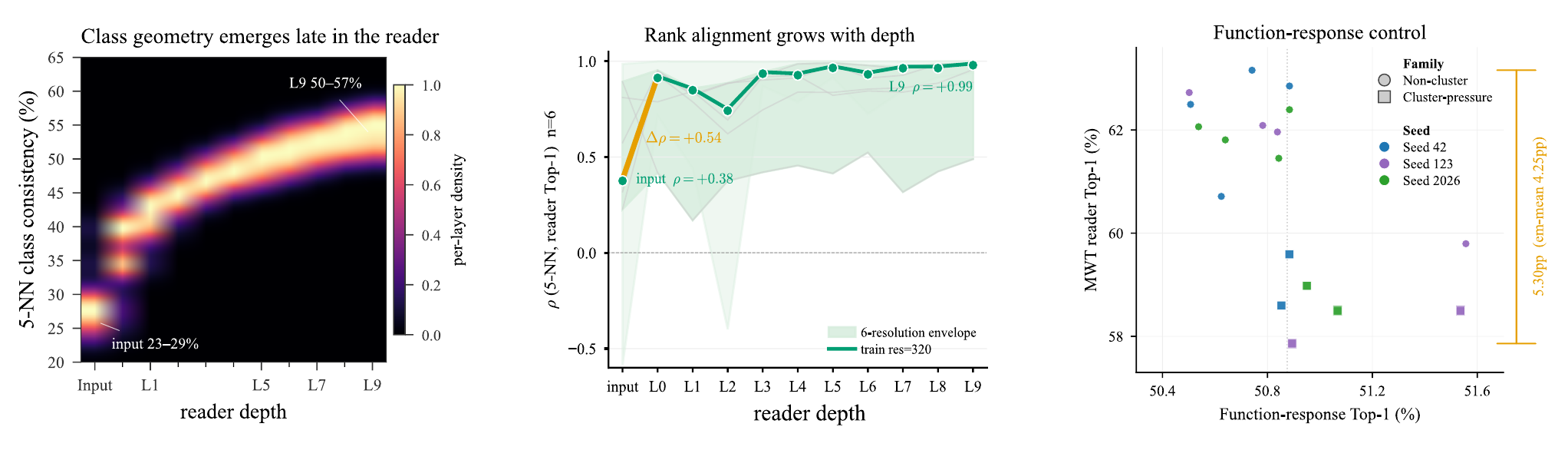}
    \vspace{-1mm}
    \caption{
    \textbf{Reader construction of class geometry and function-response control.}
    Left: token-flow 5-NN rises from the reader input to late reader states; the plotted six-emitter panel sits in the mid-20s at input and the low-to-mid-50s late in the reader.
    Center: late-state 5-NN tracks trained-reader Top-1 on the train-resolution slice, with the band showing the six-resolution envelope.
    Right: the class-only function-response reader spans only $50.51$--$50.88\%$ across emitters ($0.37$pp), far below the trained MWT weight-reader spread of $58.60$--$63.16\%$ ($4.56$pp) under checkpoint-best evaluation.
    }\label{fig:reader-construction}
    \end{figure*}

    \paragraph{Class geometry becomes readable late in the reader.}
    \figref{fig:reader-construction} traces nearest-neighbor class consistency from the reader input \(z_x\) to the reader states \(h_x^{(m)}\) defined in \eqreftext{eq:reader-state}.
    Figure~\ref{fig:reader-construction} visualizes a six-emitter panel, for which input-side 5-NN stays in the mid-20s while late reader states reach the low-to-mid-50s; \appref{app:token-flow-full} reports the exact broader-scan ranges of $21.32$--$29.48\%$ at input and $52.10$--$55.90\%$ at $h^{(9)}$.
    This localization is the important point: the reader does not merely inherit a class-aligned geometry from the native coordinate; it constructs a class-aligned representation through late token interactions.

    \paragraph{Function responses do not determine the weight-reader ranking.}
    A natural alternative is that the reader is only exploiting the realized SIREN function rather than a weight-coordinate route.
    We test this with a weight-free function-response reader that observes sparse SIREN responses at fixed coordinates and predicts the class without access to SIREN weights, following the model-free probing view of weight-space learning \citep{kahana2025probegen,horwitz2025probex}.
    The function-response reader sees nontrivial signal, but it does not reproduce the weight-reader emitter spread.
    Under the class-only setting, it has a final-window Top-1 range of $50.51$--$50.88\%$ across the six-emitter panel ($0.37$pp), whereas the matched six-configuration MWT weight reader uses checkpoint-best validation accuracy and spans $58.60$--$63.16\%$ ($4.56$pp).
    The right panel of \figref{fig:reader-construction} therefore separates two facts: sparse function responses contain class signal, but they do not determine the emitter ranking learned by the weight reader.

\begin{figure*}[t]
    \centering
    \begin{minipage}[t]{0.31\textwidth}
    \centering
    \includegraphics[width=\linewidth]{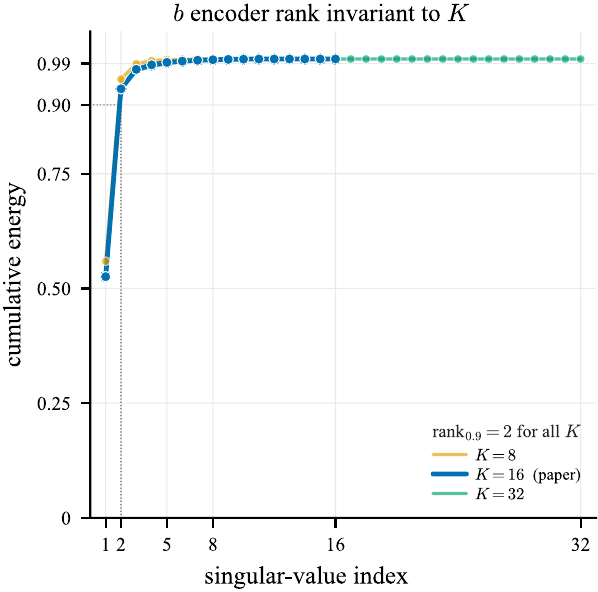}
    \end{minipage}
    \hfill
    \begin{minipage}[t]{0.31\textwidth}
    \centering
    \includegraphics[width=\linewidth]{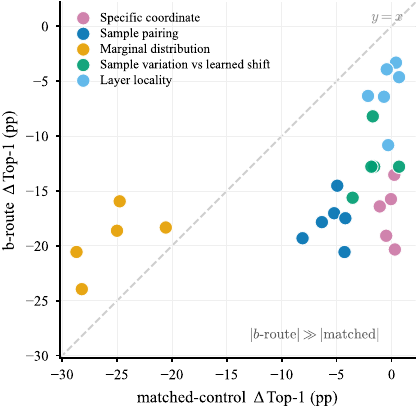}
    \end{minipage}
    \hfill
    \begin{minipage}[t]{0.31\textwidth}
    \centering
    \includegraphics[width=\linewidth]{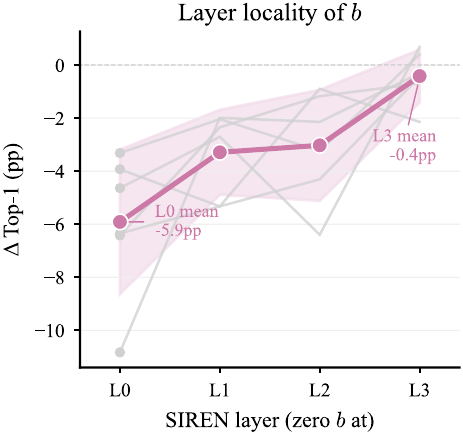}
    \end{minipage}
    \vspace{-1mm}
    \caption{
    \textbf{The SIREN bias entry is compact and causal.}
    Left: the explicit bias route has effective ranks $\mathrm{rank}_{0.9}=2$ and $\mathrm{rank}_{0.99}=5$.
    Center/right: targeted $b$-coordinate interventions are shown with matched-weight, distributional, shift-split, and layer-locality controls; \tabref{tab:b-route-controls} reports the exact ranges.
    }\label{fig:bias-column-route}
    \end{figure*}

    \paragraph{The SIREN bias entry forms a compact reader route.}
    \secref{sec:problem-formulation} defined the augmented token $t=[W,b]$.
    Here we ask whether its final SIREN bias entry is merely bookkeeping or a route used by the trained reader.
    We denote this entry by $b$, distinct from the learned coordinate shift $\beta$.
    This differs from fitting-stage INR bias mechanisms, which study how bias terms affect signal fitting or feature injection rather than reader-side weight-token readout \citep{zhang2025understandingbias}. Here $b$ is the SIREN bias entry exposed by MWT's augmented token and consumed by the trained weight reader.
    The left panel of \figref{fig:bias-column-route} gives the structural clue: in the explicit bias-route reader, the bias-encoder activations have effective rank $\mathrm{rank}_{0.9}=2$ and $\mathrm{rank}_{0.99}=5$.
    Thus the reader routes a small $b$-coordinate subspace jointly with the $W$ tokens rather than spreading the signal uniformly over token coordinates.

    \paragraph{Bias-coordinate interventions establish causality.}
    We intervene directly on the reader-visible $b$ coordinate after SIREN fitting and before the reader input projection, leaving the SIREN inner loop, the $W$ coordinates, the reader parameters, and validation images fixed.
    \tabref{tab:b-route-controls} gives the exact artifact-defense ladder, while the center and right panels of \figref{fig:bias-column-route} visualize the effect sizes and their layer locality.
    Neutralizing $b$ drops Top-1 by $13.5$--$20.3$pp, whereas a variance-matched $W$ coordinate changes Top-1 by only $-0.7$ to $+0.3$pp.
    For this split, $\delta_b$ denotes the sample-dependent residual part of the SIREN bias coordinate, and $\beta_b$ denotes the corresponding learned shift component.
    Global shuffling, dummy Gaussian or empirical replacement, and the $\delta_b/\beta_b$ split further show that the load-bearing signal is not a generic scalar column, a marginal distribution artifact, or only the learned shift.
    Removing the sample-dependent $\delta_b$ component is much more damaging than removing $\beta_b$, and layer-wise neutralization localizes the stronger effect to early SIREN bias coordinates.

    \begin{table}[t]
    \centering
    \small
    \setlength{\tabcolsep}{4.5pt}
    \renewcommand{\arraystretch}{1.1}
    \caption{
    \textbf{Bias-route causal controls.}
    $\Delta$Top-1 (pp) over applicable checkpoints on Imagenette-320 (dim$=$256).
    Mean $\pm$ sample s.d. across the applicable readers for each control axis ($n{-}1$ denominator);
    gap $= \Delta_b - \Delta_{\rm ctrl}$ (paired), with descriptive paired $t$-tests.
    }\label{tab:b-route-controls}
    \begin{tabular*}{\linewidth}{@{\extracolsep{\fill}}l@{\hspace{4pt}{\color{black!50}\vrule width 0.45pt}\hspace{4pt}}cccc}
    \toprule
    \multicolumn{1}{c}{} & \multicolumn{2}{c}{\textbf{Bias-coordinate}} & \multicolumn{2}{c}{\textbf{Matched control}} \\
    \multicolumn{1}{l}{\textbf{Control axis}} & \multicolumn{1}{c}{$\Delta_b$ mean $\pm$ sd} & \multicolumn{1}{c}{$\Delta_b$ range} & \multicolumn{1}{c}{$\Delta_{\rm ctrl}$ mean $\pm$ sd} & \multicolumn{1}{c}{gap (paired $p$)} \\
    \midrule
    Specific coordinate & $-17.0 \pm 2.7$ & $[-20.3, -13.5]$ & $-0.2 \pm 0.6$ & $\mathbf{-16.8}$ ($p{<}10^{-3}$) \\
    Sample pairing & $-17.8 \pm 2.1$ & $[-20.6, -14.5]$ & $-5.5 \pm 1.5$ & $\mathbf{-12.3}$ ($p{<}10^{-4}$) \\
    Marginal distribution & $-19.5 \pm 3.0$ & $[-23.9, -15.9]$ & $-25.4 \pm 3.3$ & $+6.0$ ($p{=}0.008$) \\
    Sample signal vs.\ learned shift & $-12.4 \pm 2.7$ & $[-15.6, -8.2]$ & $-1.6 \pm 1.5$ & $\mathbf{-10.8}$ ($p{<}10^{-3}$) \\
    Layer locality & $-5.9 \pm 2.7$ & $[-10.8, -3.3]$ & $-0.4 \pm 1.0$ & $\mathbf{-5.5}$ ($p{=}0.004$) \\
    \bottomrule
    \end{tabular*}
    \par\vspace{0.5mm}
    \begin{minipage}{\linewidth}
    \footnotesize
    Row-wise applicable-reader counts are $5$, $6$, $5$, $5$, and $6$ in table order. Rows aggregate applicable checkpoints, and architecture-specific N/A cells are excluded from row-wise $n$. The paired $t$-tests are descriptive because row-wise $n$ is small and design-defined. \appref{app:b-causal-full} reports the selected-intervention heatmap and aggregate summary used by this table; \appref{app:b-encoder-svd} reports the rank audit.
    \end{minipage}
    \end{table}

    \paragraph{Takeaway.}
    Together, these tests turn the routed account from a residual explanation into a localized mechanism.
    Class readability emerges inside the trained reader, a function-response reader does not recover the same emitter ranking, and direct $b$-coordinate interventions have large, specific effects. The next section asks whether these localized routes can be turned into controlled task-level gains.

    \section{Diagnosis-driven interventions and non-additive coupling}\label{sec:interventions}

    \tabref{tab:diagnosis-interventions} is organized across lanes to test whether the diagnosed intervention roles remain meaningful beyond the Imagenette setting; the rows are mechanism checks rather than a cross-dataset leaderboard. The clean-stack audit then asks whether the resulting gains compose independently.

    \begin{table}[t]
    \centering
    \small
    \setlength{\tabcolsep}{4.5pt}
    \renewcommand{\arraystretch}{1.1}
    \caption{
    \textbf{Diagnosis-driven interventions across three lanes.}
    Top-1 is checkpoint-best validation accuracy, reported as mean $\pm$ sample s.d. over matched seeds.
    Rows are mechanism-ordered checks rather than a leaderboard: center pressure tests exposed-coordinate pressure, routing and bias-route rows test the diagnosed reader routes, and stochastic-fit routing tests fit-density coupling.
    Rows are not fully budget-matched across lanes; \appref{app:training-hp} and \appref{app:intervention-full} record the epoch and sampling-fraction conventions.
    Bold marks the largest value within a lane only.
    }\label{tab:diagnosis-interventions}
    \begin{tabular}{l @{
      \hspace{4pt}{\color{black!50}\vrule width 0.45pt}\hspace{4pt}} cc @{
      \hspace{4pt}{\color{black!50}\vrule width 0.45pt}\hspace{4pt}} cc @{
      \hspace{4pt}{\color{black!50}\vrule width 0.45pt}\hspace{4pt}} cc}
    \toprule
    \multicolumn{1}{c}{} & \multicolumn{2}{c}{\textbf{Imagenette-320}} & \multicolumn{2}{c}{\textbf{CIFAR-10}} & \multicolumn{2}{c}{\textbf{CIFAR-10}} \\
    \multicolumn{1}{c}{} & \multicolumn{2}{c}{\scriptsize SIREN $d{=}256$, $H{\times}W{=}320$} & \multicolumn{2}{c}{\scriptsize SIREN $d{=}128$, $H{\times}W{=}32$} & \multicolumn{2}{c}{\scriptsize SIREN $d{=}256$, $H{\times}W{=}32$} \\
    \multicolumn{1}{l}{\textbf{Intervention}} &
    \multicolumn{1}{c}{Top-1 (\%)} & \multicolumn{1}{c}{$\boldsymbol{\Delta}$ (pp)} &
    \multicolumn{1}{c}{Top-1 (\%)} & \multicolumn{1}{c}{$\boldsymbol{\Delta}$ (pp)} &
    \multicolumn{1}{c}{Top-1 (\%)} & \multicolumn{1}{c}{$\boldsymbol{\Delta}$ (pp)} \\
    \midrule
    \shortstack[l]{shared-anchor\\baseline} & $60.65 \pm 0.83$ & $\phantom{+}0.00$ & $57.44 \pm 1.84$ & $\phantom{+}0.00$ & $62.16 \pm 0.05$ & $\phantom{+}0.00$ \\
    \midrule
    \shortstack[l]{center-pressure\\emitter} & $59.02 \pm 0.55$ & $-1.63$ & $55.79 \pm 0.25$ & $-1.65$ & $58.39 \pm 0.11$ & $-3.77$ \\
    \midrule
    \shortstack[l]{routing-enhanced\\reader} & $62.33 \pm 0.21$ & $+1.68$ & $59.18 \pm 0.04$ & $+1.74$ & $61.84 \pm 0.46$ & $-0.32$ \\
    \shortstack[l]{explicit bias-route\\reader} & $62.29 \pm 0.49$ & $+1.64$ & $60.70 \pm 0.29$ & $+3.26$ & $63.88 \pm 0.21$ & $+1.72$ \\
    \shortstack[l]{stochastic-fit\\routing reader} & $\mathbf{62.57 \pm 0.69}$ & $\boldsymbol{+1.92}$ & $\mathbf{64.31 \pm 0.07}$ & $\boldsymbol{+6.87}$ & $\mathbf{66.80 \pm 0.59}$ & $\boldsymbol{+4.64}$ \\
    \bottomrule
    \end{tabular}
    \par\vspace{0.5mm}
    \begin{minipage}{\linewidth}
    \footnotesize
    Center pressure is the exposed-coordinate control row.
    CIFAR-10 $d{=}256$ routing attenuation is a boundary signal, not a failed method.
    Rows are not fully budget-matched across lanes; \appref{app:intervention-full} reports per-seed values, convention checks, and composition sidecars.
    \end{minipage}
    \end{table}

    Center pressure is the exposed-coordinate control; it falls below the shared-anchor baseline in all three lanes. The route-directed rows instead act on mechanisms made visible by the diagnosis. Their lane-dependent signs are part of the stress test: routing improves Imagenette and CIFAR-10 $d{=}128$ but attenuates on CIFAR-10 $d{=}256$, while explicit bias routing and stochastic-fit routing remain positive under the lane-specific training conventions used in this panel.

    The stochastic-fit routing row changes fitting density on top of the routing-enhanced reader. It gives the largest Top-1 in this panel, but its role is diagnostic: it tests whether fit density helps a particular routed state rather than providing a portable sampling recipe.

   The clean routing-plus-bias stack also falls short of the modular null, so the non-additivity is not only a consequence of confounded bias-scale variants; \appref{app:modular-null-clean-stack} reports the exact stack audit. Likewise, the same stochastic-fit change reverses in a sampled spectral-fusion stack, showing that fit density sharpens one routed state while destabilizing another.

   Thus the interventions validate the diagnosis, while the modularity check bounds the interpretation: the gains depend on the reader state that exposes each route, rather than adding as independent modules.

    \FloatBarrier
    \section{Coordinate controls and scope}\label{sec:controls}

    One coordinate concern remains: whether the residual coordinate in \eqreftext{eq:mwt-coordinate} is merely an offset artifact. \tabref{tab:coord-packaging} addresses this concern with a reader-only control on a frozen CIFAR-10 $d{=}128$ emitter, fixing $(\theta,\beta,\lambda)$ and training fresh readers under four packagings.

    \FloatBarrier
    \begin{table}[t]
    \centering
    \small
    \setlength{\tabcolsep}{4pt}
    \renewcommand{\arraystretch}{1.1}
    \caption{
    \textbf{Coordinate-packaging control.}
    Reader-only ablation on a frozen CIFAR-10 $d{=}128$ emitter.
    Top-1 is final-5-epoch mean $\pm$ sample s.d.\ over three seeds; $\Delta_{\rm FS}$ is the gap to full+shift; id$_{95}$ is a descriptive PCA intrinsic-dimension summary. This table isolates input packaging and should not be compared numerically to the Imagenette intervention lane.
    }\label{tab:coord-packaging}
    \begin{tabular*}{\linewidth}{@{\extracolsep{\fill}}lccccc}
    \toprule
    \textbf{Packaging} & \textbf{Reader input} & $\boldsymbol{\beta}$ & $\boldsymbol{\theta}$ sub. & \textbf{id}$_{95}$ & \textbf{Top-1 / $\Delta_{\rm FS}$} \\
    \midrule
    Raw full & $\phi_x$ & -- & -- & $10.7 \pm 1.2$ & $46.17 \pm 0.73$ / $-1.14$ \\
    Full+shift & $\lambda(\phi_x + \beta)$ & \checkmark & -- & $25.7 \pm 0.9$ & $47.30 \pm 0.32$ / ref. \\
    Residual-only & $\lambda(\phi_x - \theta)$ & -- & \checkmark & $57.0 \pm 0.0$ & $43.69 \pm 0.28$ / $-3.61$ \\
    Residual+shift & $\lambda(\phi_x - \theta + \beta)$ & \checkmark & \checkmark & $55.0 \pm 0.8$ & $47.21 \pm 0.63$ / $-0.09$ \\
    \bottomrule
    \end{tabular*}
    \end{table}

    The key comparison is residual-only versus residual+shift, with full+shift as the $\beta$-including reference. Residual-only is the weak cell, whereas adding the learned shift restores performance near full+shift. Thus the residual+shift coordinate used in the paper is not an offset-only trick, but this is a packaging result rather than a claim that residual and full coordinates are intrinsically equivalent. Subtracting the anchor alone is not sufficient in this frozen-reader control. This reader-side shift $\beta$ is distinct from the SIREN bias coordinate $b$ tested in \secref{sec:routing-bias}. \appref{app:coord-packaging-raw} gives the sample/population s.d. comparison, and \appref{app:coord-packaging-full} records the nearest available boundary checks.

    Our conclusions target task-induced global SIREN weight spaces instantiated by MWT, where a shared anchor, learned update schedule, and co-trained reader shape the emitted weights. They do not rule out quotient-space, aligned, or symmetry-aware geometries in other INR regimes; they bound the native-coordinate diagnostics and reader packaging studied here. Nearest-neighbor consistency remains an exposed-geometry diagnostic, Top-1 remains the task metric, and boundary checks outside the matched pipeline are stress tests rather than additional mechanisms.

    \section{Related work}\label{sec:related-work}

    Weight-space learning asks how neural parameters can be understood, represented, or generated as structured objects \citep{eilertsen2020classifying,schurholt2022modelzoos,schurholt2022hypergenerative,peebles2022gpt,erkoc2023hyperdiffusion,han2026wsl}. Symmetry-aware readers build equivariant, attention, graph, scale, architecture, or sign/scale-aware maps over weight inputs \citep{navon2023dwsnet,zhou2023nfn,zhou2023nft,kofinas2024neuralgraphs,lim2024gmn,kalogeropoulos2024scalegmn,zhou2024unf,tran2024monomialnfn}, while model-free probing and coordinate/augmentation-based WSL read weights through functional responses or parameter tensors \citep{ashkenazi2023nern,kahana2025probegen,horwitz2025probex,shamsian2024dwsaugmentation}. We ask which induced coordinate channels a co-trained reader uses.

    Data-to-Functa and Spatial Functa learn neural fields or shift modulations as data representations \citep{dupont2022data,bauer2023spatialfuncta}; Fit-a-NeF and Implicit-Zoo study fitted neural-field datasets and downstream use \citep{papa2024fitanef,ma2024implicitzoo}; HyperINR, SPW, and ARC induce or alter semantic INR representations \citep{qiu2026hyperinr,cai2024spw,luijmes2025arc}. MWT supplies the end-to-end SIREN classification regime in which classifier feedback shapes emitted weights \citep{gielisse2025mwt}.
    Our question is not whether INR weights can support labels, but whether classifiability in such an induced coordinate is exposed as native metric geometry or made usable by the reader.

    Anchored residual coordinates are one choice within neural-weight space. Task vectors, adapters, and low-rank adaptation use reference-based updates, while permutation-invariance, re-basing, and alignment work show that raw neural coordinates are noncanonical under symmetries \citep{houlsby2019adapters,ilharco2023taskarithmetic,hu2022lora,entezari2022permutation,ainsworth2023gitrebasin,navon2024deepalign}. Neural-field modulation likewise uses shifts or latent-conditioned coordinates to package instance variation \citep{bauer2023spatialfuncta}.

    INR-bias work studies fitting-stage mechanisms; our claim concerns the downstream readout role of the SIREN \(b\)-coordinate exposed by the augmented token. Understanding Bias Terms connects INR biases to spatial aliasing and Feat-Bias feature injection \citep{zhang2025understandingbias}; ActINR, FINER, structured-dictionary analyses, and Spatial Functa relate biases or shifts to locations, frequencies, phase, or instance modulation \citep{kayabasi2025actinr,liu2024finer,yuce2022structured,bauer2023spatialfuncta}. Here \(b\) is endogenous to MWT's tokenization and consumed by a trained weight-space reader, so the explicit bias-route reader is a diagnostic of readout use rather than a fitting-stage bias-injection method.

    \section{Discussion and conclusion}\label{sec:conclusion}

    The shared-anchor SIREN system studied here gives raw geometry a favorable test: examples start from one learned point, undergo a few fitting updates, and are shaped by classifier feedback. The coordinate is not empty---its neighborhoods, probes, and packagings can move in class-consistent directions---but it is not classifier-ready. Usable class structure appears after the trained reader reorganizes weight tokens.

    The mechanism is therefore a routed readout rather than a better raw cluster.
    The SIREN bias entry supplies one compact route, but route use depends on the induced coordinate and reader state.
    In this task-induced global SIREN regime, class signal is routed rather than clustered.

    \enlargethispage{5\baselineskip}
    \paragraph{Limitations.}
    Our primary evidence relies on SIREN-based MWT and global-INR lanes; broader cross-checks provide only boundary evidence, not full task coverage. The claim is about this task-induced coordinate--reader regime, and the \(b\)-coordinate interventions test reader-side use of an emitted augmented token rather than a universal fitting-stage bias bottleneck.

{\small
\bibliographystyle{plainnat}
\bibliography{neurips_reference_camera_ready}
}

\clearpage
\appendix
\setcounter{table}{0}
\setcounter{figure}{0}
\numberwithin{table}{section}
\numberwithin{figure}{section}

\section{Model definitions and training conventions}
\label{app:source-audit}
This appendix records the model definitions and training hyperparameters used in the rest of the supplement, so later appendices can reference a single set of conventions without restating them.

\subsection{Model definitions}
\label{app:model-defs}
This subsection records the outer-loop loss conventions for the six core emitter configurations and the two single-seed-only family-panel emitters, the layer-weighted reader pool, the ten-layer MWT trunk, and the function-response reader. Symbols follow \secref{sec:problem-formulation}: $\theta$ is the shared SIREN anchor, $\phi_x=U_\alpha^k(\theta;x)$ is the image-specific SIREN parameter after $k$ inner-loop steps under the meta-learned schedule $\alpha$, $\Delta_x=\phi_x-\theta$ is the anchored residual, $z_x=\lambda(\Delta_x+\beta)$ is the reader-visible coordinate, $t_{x,\ell,j}=[W_{x,\ell,j},b_{x,\ell,j}]$ is the augmented token, and $h_x^{(m)}=R_\psi^{(m)}(z_x)$ is the reader state after the $m$-th block. Equations are reproduced from \secref{sec:problem-formulation} for cross-reference; see \eqsrefrange{eq:weights-and-residual}{eq:reader-state}.

\subsubsection{Emitter family loss conventions}
The six core configurations share the reconstruction-driven inner loop and the standard MWT outer objective, in which reconstruction and classification feedback jointly shape the shared SIREN anchor, the learned update schedule, and the trained reader. They differ only in additional auxiliary losses, reader-side route changes, or fitting-density changes. In the rows below, ``no auxiliary loss'' means no extra geometry or contrastive term beyond the standard MWT objective.
\begin{itemize}
\setlength{\itemsep}{1pt}
\item Shared-anchor baseline (Anchor): standard MWT objective, with no auxiliary geometry or contrastive loss.
\item Routing-enhanced reader (Routing): standard MWT objective with a composite reader extension; no auxiliary geometry or contrastive loss.
\item Stochastic-fit routing reader: same objective as Routing, with the inner-loop fitting density increased from $\rho = 0.05$ to $\rho = 0.10$ and the 80-epoch training-budget convention recorded in \tabref{tab:training-hp}.
\item Explicit bias-route reader: standard MWT objective with a low-rank bias encoder of width $K=16$ that compresses the $b$-coordinate of \eqreftext{eq:augmented-token} before fusion with the $W$-token stream. Reported numbers use the 80-epoch clean variant (\appref{app:b-encoder-svd}); 40-epoch ablation runs are not mixed into the 80-epoch row.
\item Center-pressure emitter (Center): the standard MWT objective plus a CenterLoss auxiliary on the raw residual coordinate $\Delta_x$, with $\lambda_{\rm center}=1.0$. This row is the raw-offset center variant; cosine-center and pre-projection CenterLoss variants are separate ablations reported only as supporting evidence.
\item Contrastive-pressure emitter (Contrast): standard MWT objective plus an InfoNCE/SupCon-style contrastive auxiliary over two augmented residual views, with temperature $\tau = 0.1$ \citep{khosla2020supcon}.
\end{itemize}

Two additional emitter configurations enter the cluster-pressure family panel of \figref{fig:geometry-gap} and \appref{app:raw-diagnostic-inventory} but are reported single-seed only:
\begin{itemize}
\setlength{\itemsep}{1pt}
\item Hierarchical cluster-pressure emitter: hierarchical gradient-clustering pressure with stage-wise grouping of residual updates over the family panel.
\item Curriculum: curriculum-style dual-head schedule with a linearly ramped supervision weight on a clustering auxiliary.
\end{itemize}
These two are reported under the family-panel convention (\secref{sec:problem-formulation}); they enter the cluster-pressure family for non-cluster-versus-cluster comparisons.

\subsubsection{Reader pool}
The shared trained reader $R_\psi$ uses a layer-weighted token pool over augmented tokens before the classification head. Let $w_m$ denote per-block attention logits over the ten reader blocks; the pooled representation is
\begin{equation}
h_x^\text{pool} = \sum_{m=1}^{10}\mathrm{softmax}(w)_m\cdot \overline{R_\psi^{(m)}(z_x)},
\label{eq:reader-pool}
\end{equation}
where $\overline{R_\psi^{(m)}(z_x)}$ is the per-token mean of the $m$-th reader block output. The pooled state is mapped to logits by a single linear head $\text{logits}_x = W_\text{out} h_x^\text{pool} + b_\text{out}$ with $W_\text{out}\in\mathbb{R}^{10\times K_\text{emb}}$. The routing-enhanced and explicit bias-route readers preserve this pooling layout and extend the per-token computation upstream of the pool.

\subsubsection{MWT trunk}
The reader trunk is a ten-layer transformer with hidden dimension $d\in\{128,256\}$ depending on the lane (Imagenette-320 uses $d=256$; CIFAR-10 lanes use $d=128$ and $d=256$ as labeled in \tabref{tab:diagnosis-interventions}). Self-attention and feed-forward blocks use the post-normalization arrangement applied to the augmented-token sequence; positional encoding follows the MWT tokenization recipe of \citet{gielisse2025mwt}.

\subsubsection{Function-response reader}
The weight-free function-response reader of \figref{fig:reader-construction} consumes sparse SIREN responses at fixed coordinates without access to SIREN weights. We evaluate four lightweight heads over the same response features:
\begin{itemize}
\setlength{\itemsep}{1pt}
\item Logreg head: linear classifier over the SIREN response values queried at the probe coordinate set.
\item kNN-5 head: $k$-nearest-neighbor classifier with $k=5$ over the same query values.
\item MLP head: two-layer multi-layer perceptron over the query values.
\item PSNR ridge head: ridge regression predicting reconstruction PSNR as an auxiliary control.
\end{itemize}
Probe-setting variations (random-256 versus 32$\times$32 grid coordinates, with optional reconstruction-auxiliary weight $w_{\rm psnr}\in\{0,1,10\}$) appear in \appref{app:function-response-full}. The function-response reader is trained for 30 epochs on a frozen emitter and reported as final-window Top-1 over the last five epochs.
\FloatBarrier

\subsection{Training hyperparameters}
\label{app:training-hp}
Training hyperparameters for the six core emitter configurations on the locked Imagenette-320, $d=256$ lane are summarized in \tabref{tab:training-hp}. All configurations use AdamW with weight decay $10^{-4}$ and OneCycleLR with $5\%$ warmup; reader-side learning rates are $1\times10^{-4}$ for the SIREN and classifier parameter groups and $1\times10^{-2}$ for the MetaSGD inner-loop learning rates. CIFAR-10 lanes inherit the same outer optimizer with epoch and sampling fractions matched to each lane, as summarized in \appref{app:intervention-sidecar}.

\begin{table}[!htbp]
\centering
\small
\caption{Per-configuration training hyperparameters on the Imagenette-320, $d=256$ lane. All rows share AdamW with weight decay $10^{-4}$ and OneCycleLR with $5\%$ warmup. The explicit bias-route reader row uses the epoch-matched 80-epoch variant rather than 40-epoch ablation runs. Configuration names follow \appref{app:model-defs}.}
\label{tab:training-hp}
\begin{tabular}{lccccl}
\toprule
Configuration & epochs & batch size & inner steps & sample fraction $\rho$ & loss-specific \\
\midrule
Anchor & 40 & 16 & 4 & 0.05 & --- \\
Routing & 40 & 16 & 4 & 0.05 & --- \\
\shortstack[l]{stochastic-fit\\routing reader} & 80 & 16 & 4 & 0.10 & --- \\
\shortstack[l]{explicit bias-route\\reader} & 80 & 16 & 4 & 0.05 & --- \\
Center & 40 & 16 & 4 & 0.05 & $\lambda_\text{center}=1.0$ \\
Contrast & 40 & 16 & 4 & 0.05 & $\tau=0.1$ \\
\bottomrule
\end{tabular}
\end{table}
\FloatBarrier

\section{Raw-coordinate diagnostics}
\label{app:raw-diagnostic-inventory}
This section supports the negative raw-geometry result in \secref{sec:geometry-gap} by tabulating the full diagnostic inventory across local 5-NN, component summaries, spectral/effective-rank summaries, Fisher-style summaries, intrinsic/global geometry, and reader-input projection spectra. The claim is bounded: these native-coordinate diagnostics do not monotonically explain trained-reader accuracy, but we do not rule out quotient-space, aligned, or symmetry-aware geometries outside the coordinate studied here.

The family panel partitions emitter configurations into a non-cluster group ($n_\text{NC}=12$ routing-variant configurations) and a cluster-pressure group ($n_\text{CL}=4$, comprising the Center, Contrast, hierarchical cluster-pressure emitter, and Curriculum configurations). Per-configuration loss conventions are recorded in \appref{app:model-defs}; per-configuration training hyperparameters are in \appref{app:training-hp}. Per-emitter heterogeneity inside the cluster-pressure group is reported in \appref{app:geometry-heterogeneity}.

\subsection{Raw-coordinate diagnostic matrix}

\begin{table}[!htbp]
\centering
\small
\caption{Raw-coordinate diagnostic matrix. Values are family means $\pm$ sample s.d.\ across emitter configurations within each family ($n_\text{NC}=12$ non-cluster, $n_\text{CL}=4$ cluster-pressure; family-level claim, not per-emitter theorem). The first row is the trained-reader task metric; subsequent rows are diagnostics on the exposed coordinate or its components. Cluster-pressure sample s.d.\ is much larger than non-cluster sample s.d.\ on local 5-NN and effective-rank rows; per-emitter heterogeneity inside the cluster-pressure family is plotted in \appref{app:geometry-heterogeneity}.}
\label{tab:raw-diagnostic-matrix}
\begin{tabular*}{\textwidth}{@{\extracolsep{\fill}}lccc}
\toprule
Quantity & NC ($n=12$) & CL ($n=4$) & $\Delta$ \\
\midrule
Trained-reader Top-1 (\%) & 61.65 $\pm$ 0.45 & 59.24 $\pm$ 0.62 & $-2.41$ pp \\
Full-offset 5-NN (\%) & 23.75 $\pm$ 0.66 & 25.70 $\pm$ 3.89 & $+1.95$ pp \\
W-only 5-NN (\%) & 22.32 $\pm$ 0.42 & 24.73 $\pm$ 3.48 & $+2.41$ pp \\
All-bias 5-NN (\%) & 28.76 $\pm$ 1.08 & 28.43 $\pm$ 1.11 & $-0.33$ pp \\
W3 effective rank & 130.5 $\pm$ 4.9 & 104.2 $\pm$ 49.7 & $-26.3$ \\
W4 effective rank & 125.8 $\pm$ 5.1 & 112.5 $\pm$ 42.8 & $-13.2$ \\
Bias Fisher mean & 0.1090 $\pm$ 0.0019 & 0.1088 $\pm$ 0.0049 & $-0.0002$ \\
Bias Fisher top-50 5-NN (\%) & 27.14 $\pm$ 1.28 & 26.58 $\pm$ 0.40 & $-0.57$ pp \\
\bottomrule
\end{tabular*}
\end{table}

\FloatBarrier

The first three rows reproduce the sign reversal used in \figref{fig:geometry-gap}: local class consistency rises under cluster pressure, while the trained reader falls. The remaining rows broaden the check beyond a single 5-NN score. Some component and spectral summaries move, but not in a way that restores a monotonic explanation of validation accuracy. The cluster-pressure sample s.d.\ on the 5-NN and effective-rank rows is three to ten times larger than the non-cluster sample s.d., reflecting per-emitter heterogeneity inside the cluster-pressure family rather than a per-emitter theorem; we therefore frame this matrix as a family-level diagnostic, not a per-emitter guarantee.
\FloatBarrier
\section{Geometry-gap robustness}
\label{app:geometry-heterogeneity}
This appendix keeps the family-level sign reversal in \figref{fig:geometry-gap} honest by reporting the matched six-emitter geometry audit at per-configuration resolution under cross-seed averaging. The main figure is a family-level stress test, not a per-emitter theorem; the audit below shows how standard raw and late-reader geometry cells behave under the same cluster-versus-non-cluster comparison after pooling three seeds per emitter. This six-emitter sub-panel is complementary to the input-side probe view of \figref{fig:input-side-no-rescue}: that figure uses a learnable probe (Logreg, $k$NN-5, MLP), whereas the table below isolates a non-parametric geometry view (5-NN class consistency and nearest-centroid Top-1) on the same paired panel; the two views move together at the late-reader state and disagree at the raw and input-projection states.

\subsection{Paired six-emitter geometry audit}
\begin{table}[!htbp]
\centering
\small
\caption{Six-emitter paired-geometry audit, cross-seed pivot. Per-configuration cells are mean $\pm$ sample s.d.\ over three seeds (seeds 42/123/2026). Family means and family sample s.d.\ are taken across configurations within each family ($n_\text{NC}=4$ non-cluster: Anchor, Routing, stochastic-fit routing reader, explicit bias-route reader; $n_\text{CL}=2$ cluster-pressure: Center, Contrast). $\Delta$ is the family-mean difference (cluster $-$ non-cluster, percentage points); the rightmost column reports a Welch unequal-variance $t$ statistic with Welch--Satterthwaite degrees of freedom. Cluster-pressure family s.d.\ on the late-reader 5-NN row is small here because per-emitter values are tightly clustered around 53--54\% across both configurations. Raw residual rows do not reproduce a sign reversal at the six-emitter sub-panel; the family-level reversal is restored on the broader 16-emitter panel of \figref{fig:geometry-gap} (full-offset 5-NN $+1.95$pp, $n=12$ vs $n=4$).}
\label{tab:paired-geometry-audit}
\begin{tabular}{lcccc}
\toprule
Metric & NC ($n=4$) & CL ($n=2$) & $\Delta$ & Welch $t$ (df) \\
\midrule
5-NN, raw residual PCA-128 (\%) & $20.70 \pm 0.63$ & $20.36 \pm 1.61$ & $-0.35$ & $-0.29$ (1.2) \\
Nearest-centroid Top-1, raw PCA-128 (\%) & $41.38 \pm 0.47$ & $41.41 \pm 0.46$ & $+0.03$ & $+0.06$ (2.1) \\
5-NN, input projection $z^0$ (\%) & $28.71 \pm 0.96$ & $27.27 \pm 3.27$ & $-1.44$ & $-0.61$ (1.1) \\
5-NN, reader block $h^{(9)}$ (\%) & $58.61 \pm 2.01$ & $53.68 \pm 0.46$ & $\mathbf{-4.94}$ & $\mathbf{-4.67}$ (3.5) \\
Nearest-centroid, reader block $h^{(9)}$ (\%) & $59.68 \pm 1.17$ & $55.76 \pm 0.50$ & $\mathbf{-3.92}$ & $\mathbf{-5.73}$ (4.0) \\
\bottomrule
\end{tabular}
\end{table}
\FloatBarrier

The two late-reader rows reproduce the sign reversal expected under the routed account: cluster-pressure configurations sit roughly $4$--$5$pp below non-cluster configurations on $h^{(9)}$ neighborhood metrics. The raw residual PCA-128 row is essentially neutral on this six-emitter sub-panel ($\Delta = -0.35$pp, Welch $t = -0.29$); the broader 16-emitter family panel of \figref{fig:geometry-gap} carries a different sign at full-offset 5-NN ($+1.95$pp, $n_\text{NC}=12$, $n_\text{CL}=4$). The two panels are not contradictory: the 16-emitter panel mixes 12 routing-variant non-cluster configurations and is the family-level stress test, whereas the table above uses only the four non-cluster configurations that have full three-seed coverage and is therefore underpowered on the raw side.

\section{Input-side, reconstruction, and fresh-reader controls}
\label{app:input-readouts-full}
This section tests whether a different shallow readout, a reconstruction-quality target, or an independently trained reader rescues the clustered account. The controls operate in two lanes: co-trained reader probes for input-side and late-reader states, and fresh-reader controls on frozen emitters. Their absolute accuracies are not directly comparable, but both lanes test whether cluster-pressure emitters become easier for the reader.

\subsection{Exact input-side and late-reader readout gaps}
\begin{table}[!htbp]
\centering
\small
\caption{Input-side and late-reader readout contrasts, cross-seed pivot. Values are family means over the six-configuration pivot; per-configuration cells are three-seed means. $\Delta$ is cluster-pressure minus non-cluster, and the right column reports Welch $t$ statistics. Shallow input-side probes remain mixed and small, while late-reader probes move in the trained-reader direction.}
\label{tab:input-readouts-full}
\begin{tabular}{lcccc}
\toprule
Metric & NC ($n=4$) & CL ($n=2$) & $\Delta$ & Welch $t$ (df) \\
\midrule
Class Logreg, raw PCA-128 (\%) & $40.15 \pm 0.40$ & $39.47 \pm 0.93$ & $-0.68$ pp & $-0.99$ (1.2) \\
Class 5-NN, raw PCA-128 (\%) & $17.70 \pm 0.51$ & $17.88 \pm 0.36$ & $+0.18$ pp & $+0.50$ (3.0) \\
PSNR ridge $R^2$, raw PCA-128 & $0.047 \pm 0.025$ & $0.057 \pm 0.056$ & $+0.010$ & $+0.25$ (1.2) \\
Class Logreg, $z^0$ projection (\%) & $36.22 \pm 1.03$ & $37.01 \pm 1.71$ & $+0.79$ pp & $+0.60$ (1.4) \\
Class 5-NN, $z^0$ projection (\%) & $26.79 \pm 1.31$ & $25.46 \pm 1.65$ & $-1.34$ pp & $-1.00$ (1.7) \\
PSNR ridge $R^2$, $z^0$ projection & $0.058 \pm 0.026$ & $0.037 \pm 0.015$ & $-0.021$ & $-1.21$ (3.5) \\
Class Logreg, $h^{(9)}$ (\%) & $64.12 \pm 1.68$ & $60.34 \pm 0.84$ & $\mathbf{-3.78}$ pp & $\mathbf{-3.67}$ (3.9) \\
Class 5-NN, $h^{(9)}$ (\%) & $56.69 \pm 0.80$ & $52.00 \pm 1.65$ & $\mathbf{-4.69}$ pp & $\mathbf{-3.80}$ (1.2) \\
PSNR ridge $R^2$, $h^{(9)}$ & $0.705 \pm 0.052$ & $0.650 \pm 0.159$ & $-0.055$ & $-0.48$ (1.1) \\
\bottomrule
\end{tabular}
\end{table}
\FloatBarrier

\subsection{Fresh-reader controls on frozen emitters}
\label{app:seed-stability}
\begin{table}[!htbp]
\centering
\small
\caption{Fresh-reader controls on frozen emitters (seed-stability check). Values are final-window Top-1 percentages for independently trained fresh readers; the seed order is 42 / 123 / 2026. These 50--54\% controls operate in the fresh-reader lane and should not be compared numerically to co-trained 60--63\% intervention rows. The $\Delta$ contrast is paired across the three matched seeds and yields a sign-stable negative $\Delta$ in all three seeds, with cluster-pressure family s.d.\ smaller than non-cluster family s.d.\ ($0.32$ vs $1.47$pp); cluster-pressure configurations are tightly clustered around 51--52\% rather than spreading further apart. Configuration names follow the abbreviations defined in \appref{app:model-defs}.}
\label{tab:fresh-reader-pivot}
\begin{tabular}{lccc}
\toprule
Configuration & mean (\%) & sample s.d. (pp) & role \\
\midrule
Anchor & 50.61 & 0.48 & non-cluster \\
Routing & 52.91 & 0.41 & non-cluster \\
\shortstack[l]{stochastic-fit\\routing reader} & 54.30 & 0.49 & non-cluster \\
\shortstack[l]{explicit bias-route\\reader} & 53.46 & 0.11 & non-cluster \\
Center & 51.85 & 0.39 & cluster-pressure \\
Contrast & 51.68 & 0.30 & cluster-pressure \\
\midrule
NC mean ($n=4$) & 52.82 & 1.47 & group \\
CL mean ($n=2$) & 51.77 & 0.32 & group \\
$\Delta$ & $-1.05$ & --- & paired $t = -2.37$, $\mathrm{df}=2$, $p \approx 0.14$ \\
\bottomrule
\end{tabular}
\end{table}
\FloatBarrier

\section{Token-flow routing diagnostics}
\label{app:token-flow-full}
This appendix documents the reader-depth diagnostic used in \secref{sec:routing-bias}. The figure below uses a broader 19-configuration token-flow scan as a family-level robustness check, while the per-configuration table reports the 9-configuration sub-panel used for the main reader-depth claim. The main-text Figure~\ref{fig:reader-construction} uses a separate six-emitter visualization panel, so the appendix reports exact broad-scan ranges rather than the rounded Figure~\ref{fig:reader-construction} ranges. The token-flow curve is the mechanism evidence; cross-resolution architecture comparisons are omitted here because they are boundary checks rather than reader-depth mechanism evidence. The appendix is therefore descriptive rather than exhaustive: the broad scan shows a representative run, and the 9-configuration table localizes where class geometry becomes readable.

\begin{figure}[!htbp]
\centering
\includegraphics[width=\textwidth]{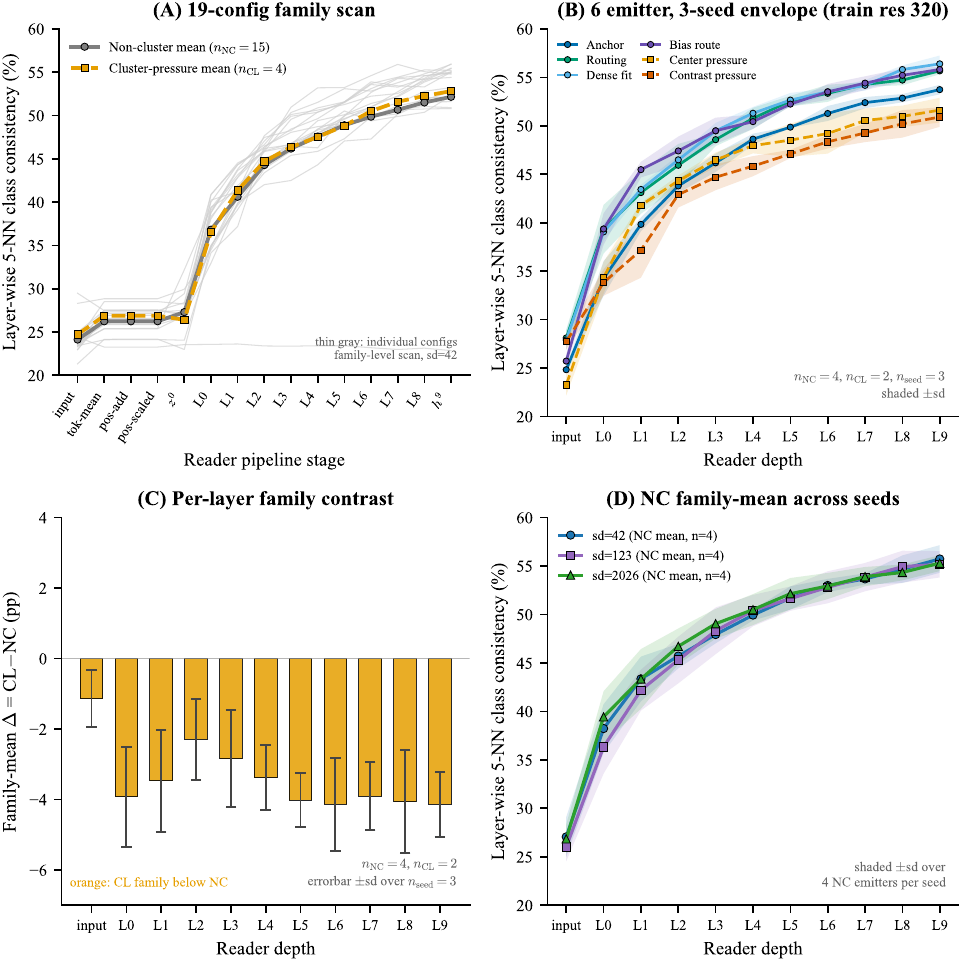}
\caption{Layer-wise token-flow curve over the 19-configuration scan ($n_\text{NC}=15$ non-cluster routing variants, $n_\text{CL}=4$ cluster-pressure configurations). Thin gray lines show individual configurations; the thick blue line shows the non-cluster family mean and the thick dashed orange line shows the cluster-pressure family mean. Class-neighborhood consistency becomes readable inside late reader states. Stage-wise 5-NN ranges and correlations against trained-reader Top-1 are: input raw offset, range 21.32--29.48\%, Pearson $r=-0.34$, Spearman $\rho=0.10$; reader input projection $z^0$, range 22.98--29.43\%, $r=0.59$, $\rho=0.68$; reader block $h^{(4)}$, range 45.86--51.52\%, $r=0.68$, $\rho=0.63$; reader block $h^{(9)}$, range 52.10--55.90\%, $r=0.94$, $\rho=0.65$. These exact ranges are for the broader 19-configuration scan and therefore need not numerically match the rounded six-emitter summary in Figure~\ref{fig:reader-construction}.}
\label{fig:app-token-flow-curve}
\end{figure}

The paper uses panel sizes that differ by section: \figref{fig:geometry-gap} reports a 16-configuration family panel (12 non-cluster routing variants and 4 cluster-pressure configurations); \figsrefrange{fig:input-side-no-rescue}{fig:reader-construction} use a 6-configuration sub-panel (the four core non-cluster configurations and the two cluster-pressure configurations); \appref{app:input-readouts-full} uses the same 6-configuration sub-panel; \figref{fig:app-token-flow-curve} below uses a broader 19-configuration scan; and \tabref{tab:token-flow-per-emitter} uses a 9-configuration sub-panel that combines the four core paper configurations with five additional routing variants. These different panel sizes serve different roles: the 16- and 19-configuration scans test family-level robustness, the 6-configuration sub-panel anchors per-configuration reproducibility, and the 9-configuration sub-panel localizes where class geometry becomes readable.

\subsection{Layer-wise 5-NN curve and correlation}
\FloatBarrier

\subsection{Selected per-configuration token-flow values}

\begin{table}[!htbp]
\centering
\small
\caption{Selected token-flow values for the 9-configuration sub-panel shown in \figref{fig:app-token-flow-curve}. The four core paper rows are Anchor, Routing, Center, and Contrast; the remaining five rows are auxiliary routing variants included to show where late-reader class geometry becomes readable.}
\label{tab:token-flow-per-emitter}
\setlength{\tabcolsep}{4pt}
\begin{tabular*}{\textwidth}{@{\extracolsep{\fill}}p{0.30\textwidth}cccc}
\toprule
Configuration & Top-1 (\%) & input 5-NN (\%) & $z^0$ projection 5-NN (\%) & $h^{(9)}$ 5-NN (\%) \\
\midrule
Anchor baseline & 60.71 & 23.39 & 26.65 & 53.22 \\
Dual-attention reader & 62.11 & 23.54 & 27.11 & 55.01 \\
Progressive-feedback reader & 62.06 & 25.04 & 27.57 & 55.41 \\
Layer-weighted reader & 61.61 & 23.80 & 26.75 & 55.03 \\
Contrast emitter & 58.60 & 29.48 & 26.90 & 52.10 \\
Center emitter & 59.59 & 21.32 & 22.98 & 52.79 \\
Bias-scaled reader & 62.17 & 23.57 & 27.11 & 55.03 \\
Two-component routing stack & 61.91 & 24.87 & 29.43 & 55.90 \\
Routing-enhanced reader & 62.50 & 24.97 & 28.38 & 55.21 \\
\bottomrule
\end{tabular*}
\end{table}

\FloatBarrier

\section{Function-response and alternative-reader controls}
\label{app:function-response-full}
This section tests whether the trained weight reader is merely recovering class signal from SIREN function responses. The function-response reader queries emitted SIRENs as functions instead of consuming augmented weight tokens; located alternative-reader rows are boundary checks, not the main mechanism evidence. On the six-configuration panel used for the Figure~\ref{fig:reader-construction} right-panel comparison, the trained-reader checkpoint-best Top-1 values are 60.71 (Anchor), 62.50 (Routing), 63.16 (stochastic-fit routing), 62.85 (explicit bias-route), 59.59 (Center), and 58.60 (Contrast), giving a spread of 58.60--63.16\%; this spread is not numerically comparable to the function-response final-window spread because the evaluation conventions differ.

\subsection{Exact six-configuration function-response values}
\begin{table}[!htbp]
\centering
\small
\caption{Trained-reader versus function-response values, cross-seed pivot. Values are Top-1 percentages; $\Delta$ is cluster-pressure minus non-cluster. The trained reader uses checkpoint-best validation accuracy, while function-response rows use random-256 responses and final-window accuracy with optional PSNR auxiliary weight $w_{\rm psnr}$. Function-response gaps remain flat relative to the trained-reader gap.}
\label{tab:function-response-exact-values}
\begin{tabular}{lcccc}
\toprule
Quantity & NC ($n=4$) & CL ($n=2$) & $\Delta$ & Welch $t$ (df) \\
\midrule
Trained reader & $61.96 \pm 0.88$ & $58.67 \pm 0.50$ & $\mathbf{-3.29}$ pp & $\mathbf{-5.84}$ (3.6) \\
Function-response, $w_{\rm psnr}=0$ & $50.78 \pm 0.16$ & $51.03 \pm 0.13$ & $+0.25$ pp & $+2.05$ (2.6) \\
Function-response, $w_{\rm psnr}=1$ & $47.24 \pm 0.20$ & $47.37 \pm 0.06$ & $+0.13$ pp & $+1.19$ (3.8) \\
Function-response, $w_{\rm psnr}=10$ & $32.50 \pm 0.20$ & $32.71 \pm 0.49$ & $+0.21$ pp & $+0.57$ (1.2) \\
\bottomrule
\end{tabular}
\end{table}
\FloatBarrier

\subsection{Probe-setting sweep}
\begin{table}[!htbp]
\centering
\small
\caption{Function-response probe-setting sweep. Rows vary the query set and PSNR auxiliary weight $w_{\rm psnr}\in\{0,1,10\}$. Random-256 rows use cross-seed averages when available; $32\times32$ grid rows are seed-42 single-seed checks. Values are Top-1 ranges and configuration-level means over the six configurations.}
\label{tab:function-response-sweep}
\begin{tabular}{lccc}
\toprule
Setting & range (\%) & mean (\%) & s.d.\ (pp) \\
\midrule
random-256, $w_{\rm psnr}=0$ & 50.51--50.88 & 50.75 & 0.16 \\
random-256, $w_{\rm psnr}=1$ & 46.93--47.47 & 47.21 & 0.18 \\
random-256, $w_{\rm psnr}=10$ & 32.44--33.17 & 32.69 & 0.27 \\
32$\times$32 grid, $w_{\rm psnr}=0$ & 49.75--50.44 & 50.05 & 0.25 \\
32$\times$32 grid, $w_{\rm psnr}=1$ & 47.53--48.62 & 48.03 & 0.39 \\
32$\times$32 grid, $w_{\rm psnr}=10$ & 30.82--31.28 & 31.09 & 0.21 \\
\bottomrule
\end{tabular}
\end{table}
\FloatBarrier

\subsection{Alternative-reader boundary table}
The alternative-reader rows below are bounded boundary checks rather than direct emitter swaps. Three caveats apply when reading these numbers: (i) the DWSNet row uses a short 10-epoch schedule on a frozen CIFAR-10 $d=128$ weight set, so the 31.47\% should be read as a scale check rather than a converged DWSNet score; (ii) the Track-2 rows use a 5k-SIREN dataset rather than weights emitted by the MWT pipeline, so they are best read as the closest available proxy for a reader swap rather than a literal swap; (iii) the NFT rows report best-epoch validation Top-1, and their sizable best-versus-final gap means these values should be read as an upper bound rather than a stable end-of-training operating point.

\begin{table}[!htbp]
\centering
\small
\caption{Located alternative-reader stress tests. Best Top-1 in percent. The DWS-style row uses a short 10-epoch schedule on frozen CIFAR-10 $d=128$ weights; the MWT and NFT rows use the Track-2 5k SIREN set as the closest available proxy for a reader swap on frozen MWT-style emitters; the NFT row reports best-epoch validation Top-1 for the original and paper-default schedules. These are boundary checks on reader conventions, not main mechanism evidence; the three caveats in the prose above bound their interpretation.}
\label{tab:alternative-reader-boundary}
\begin{tabular}{lc}
\toprule
Alternative-reader setting & Best Top-1 (\%) \\
\midrule
DWS-style, CIFAR-10 $d=128$ & 31.47 \\
MWT, Track-2 & 13.20 \\
NFT, Track-2 (original / paper-default) & 16.50 / 19.20 \\
\bottomrule
\end{tabular}
\end{table}
\FloatBarrier

\section{Bias-coordinate causal controls}
\label{app:b-causal-full}
This appendix expands the bias-route causal controls used in \secref{sec:routing-bias}. It addresses arbitrary-scalar, distributional, and shift-only explanations.
We report a heatmap over selected interventions and a compact matrix summary. Each intervention acts on the reader-visible SIREN bias coordinate or on a matched weight-coordinate control while leaving validation images fixed.
The pattern is specificity: bias-coordinate interventions produce large drops, while matched controls are materially smaller and vary by control family. Architecture-specific N/A cells are treated as not applicable, not as failed runs.

\subsection{Causal ladder matrix}
\begin{figure}[!htbp]
\centering
\includegraphics[width=\textwidth]{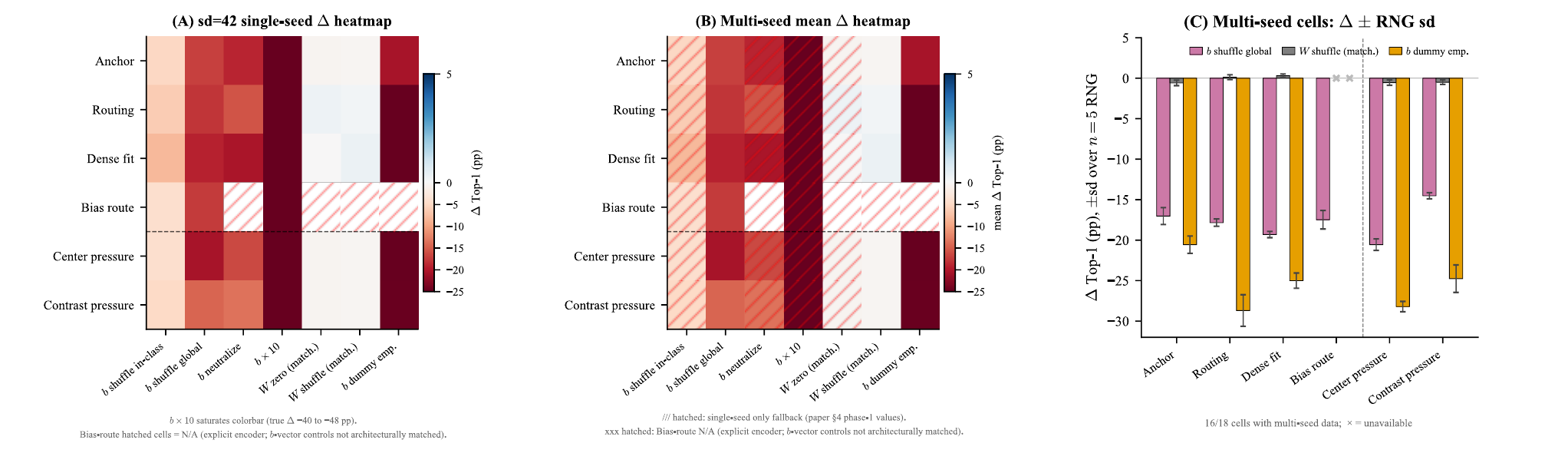}
\caption{Selected bias-coordinate interventions across configurations. Entries are Top-1 changes relative to each configuration's baseline. The heatmap visualizes targeted $b$-coordinate effects alongside matched-weight, distributional, shift-split, and layer-locality controls; \tabref{tab:b-route-controls} reports the exact ranges.}
\label{fig:app-b-causal-heatmap}
\end{figure}
\FloatBarrier

\subsection{Aggregate intervention summary}

\begin{table}[!htbp]
\centering
\small
\caption{Bias-coordinate intervention summary. Values are mean Top-1 changes relative to each configuration's baseline; ranges are min--max over applicable configurations. Rows group specificity, sample pairing, distributional dummy, sample-variation-versus-shift, and layer-locality controls. N/A entries occur only when the explicit bias-route reader changes the pre-projection structure. Rows labeled ``keep'' report the retained component; the parenthetical states which component is removed.}
\label{tab:b-causal-summary}
\setlength{\tabcolsep}{4pt}
\begin{tabular*}{\textwidth}{@{\extracolsep{\fill}}p{0.38\textwidth}ccc}
\toprule
Intervention axis & applicable configurations & mean $\Delta$ (pp) & range (pp) \\
\midrule
Bias neutralize & 5 & $-17.02$ & $-20.33$ to $-13.53$ \\
Matched-weight neutralize & 5 & $-0.35$ & $-0.69$ to $+0.25$ \\
Cross-sample shuffle & 6 & $-17.79$ & $-20.57$ to $-14.52$ \\
Within-class shuffle & 6 & $-5.52$ & $-8.10$ to $-4.20$ \\
Gaussian dummy & 5 & $-19.48$ & $-23.94$ to $-15.94$ \\
Empirical dummy & 5 & $-25.45$ & $-28.69$ to $-20.56$ \\
Keep $\delta_b$ only (remove $\beta_b$) & 5 & $-1.59$ & $-3.54$ to $+0.69$ \\
Keep $\beta_b$ only (remove $\delta_b$) & 5 & $-12.44$ & $-15.62$ to $-8.20$ \\
Layer-0 only & 6 & $-5.91$ & $-10.83$ to $-3.31$ \\
Layer-3 only & 6 & $-0.41$ & $-2.14$ to $+0.69$ \\
\bottomrule
\end{tabular*}
\end{table}

\FloatBarrier

\section{Bias-encoder spectra}
\label{app:b-encoder-svd}
This section supports the compact-route claim and prevents confusion between the bias-encoder width $K$ and the nearest-neighbor $k$ in 5-NN diagnostics. The spectra show that most energy is captured by two directions, while the 0.99-energy threshold uses about four to five dimensions depending on encoder width. The rank table is a threshold audit, not a method search.

The bias-encoder rank claim is robust along three independent axes. The cross-$K$ axis is reported in this appendix: across $K\in\{8,16,32\}$, the 0.9-energy effective rank is fixed at 2 and the 0.99-energy effective rank is 4--5 (\tabref{tab:b-encoder-k-sweep}). The cross-architecture axis combines the bias-encoder-only variant numbers reported in main \figrefshort{fig:bias-column-route} (left) with the explicit bias-route reader numbers used in the intervention table (\secref{sec:interventions}); both architectures hit the same integer rank at $K=16$ ($\mathrm{rank}_{0.9}=2$, $\mathrm{rank}_{0.99}=5$). The cross-seed axis was spot-checked at $K=16$ across three seeds and gave the same integer rank in the available spectra, but that auxiliary check is not tabulated separately here. Together these three robustness axes mean the rank-2 finding does not depend on encoder width, encoder architecture variant, or the available seed checks.

\subsection{Singular spectrum and cumulative energy}
\begin{figure}[!htbp]
\centering
\includegraphics[width=\textwidth]{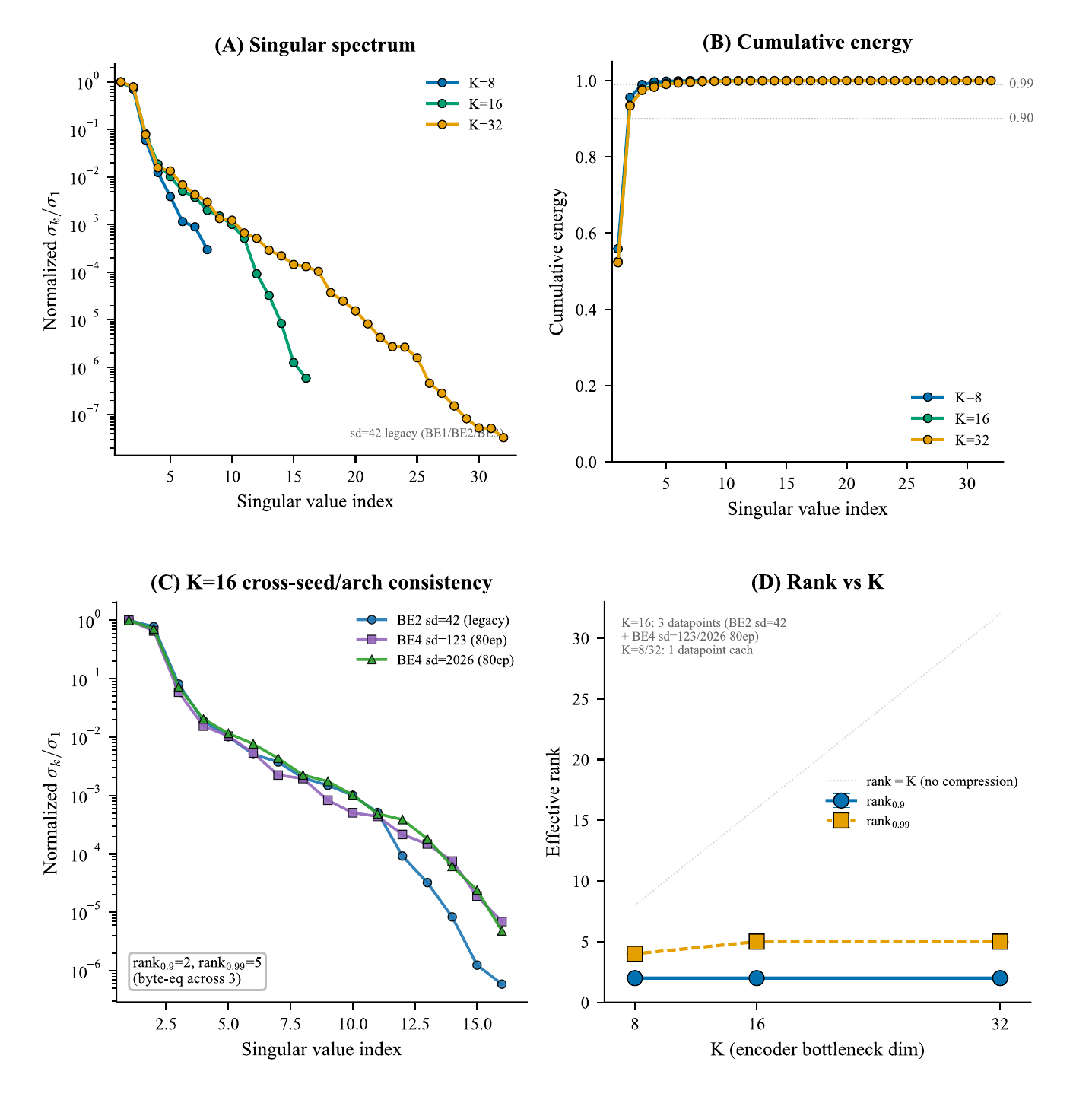}
\caption{Bias-encoder singular spectra (cross-$K$ view, complementary to main \figrefshort{fig:bias-column-route} left which fixes $K=16$ and reports cross-seed explicit bias-route reader spectra). Solid lines show normalized singular values; dashed lines show cumulative energy. Horizontal reference lines mark 0.90 and 0.99 cumulative energy.}
\label{fig:app-b-encoder-spectrum}
\end{figure}
\FloatBarrier

\subsection{Rank table}

\begin{table}[!htbp]
\centering
\small
\caption{Bias-encoder SVD rank across encoder widths (cross-$K$ axis; cross-architecture and the auxiliary cross-seed spot-check are summarized in main \figrefshort{fig:bias-column-route} left and in the section preamble above). Here $K$ is the bias-encoder width, not the nearest-neighbor $k$ used in 5-NN diagnostics. Effective rank uses cumulative spectral energy. The $K=8$ and $K=32$ rows summarize the corresponding width ablations, while the $K=16$ row is additionally spot-checked across three seeds without a separate table. The rank-2 finding is therefore stable across the reported robustness axes.}
\label{tab:b-encoder-k-sweep}
\begin{tabular*}{\textwidth}{@{\extracolsep{\fill}}lccccc}
\toprule
Encoder variant & $K$ & Top-1 (\%) & $\mathrm{rank}_{0.9}$ & $\mathrm{rank}_{0.99}$ & top-2 energy (\%) \\
\midrule
bias encoder & 8 & 61.50 & 2 & 4 & 95.6 \\
bias encoder & 16 & 61.40 & 2 & 5 & 93.5 \\
bias encoder & 32 & 60.92 & 2 & 5 & 93.4 \\
\bottomrule
\end{tabular*}
\end{table}

\FloatBarrier

\section{Intervention lanes and modularity}
\label{app:intervention-sidecar}
\label{app:intervention-full}
\label{app:intervention-convention-audit}
This appendix supports \secref{sec:interventions} by attaching per-seed values, convention notes, and composition checks to the main intervention panel. It keeps the mechanism interpretation separate from the main text: center pressure stays below the anchor, routing alone is lane-dependent, and explicit bias-route and stochastic-fit rows remain positive under the lane-specific conventions used in this panel.

\subsection{Main intervention lane across three datasets}
\begin{table}[!htbp]
\centering
\small
\caption{Main intervention values across three lanes. Values are checkpoint-best Top-1, reported as mean $\pm$ sample s.d. over matched seeds. The rows repeat the mechanism-ordered main panel to show per-lane seed variability and convention notes. Rows are not fully budget-matched; epoch and sampling-fraction conventions are recorded in \appref{app:training-hp}.}
\label{tab:main-intervention-cross-dataset}
\begin{tabular}{lccc}
\toprule
Intervention & Imagenette ($d=256$) & CIFAR-10 ($d=128$) & CIFAR-10 ($d=256$) \\
\midrule
shared-anchor baseline & $60.65 \pm 0.83$ & $57.44 \pm 1.84$ & $62.16 \pm 0.05$ \\
center-pressure emitter & $59.02 \pm 0.55$ & $55.79 \pm 0.25$ & $58.39 \pm 0.11$ \\
routing-enhanced reader & $62.33 \pm 0.21$ & $59.18 \pm 0.04$ & $61.84 \pm 0.46$ \\
explicit bias-route reader & $62.29 \pm 0.49$ & $60.70 \pm 0.29$ & $63.88 \pm 0.21$ \\
stochastic-fit routing reader & $62.57 \pm 0.69$ & $\mathbf{64.31 \pm 0.07}$ & $\mathbf{66.80 \pm 0.59}$ \\
\bottomrule
\end{tabular}
\end{table}
\FloatBarrier

\subsection{Cross-dataset 6-fork sidecar (Imagenette + CIFAR-10 lanes)}
\label{app:cifar-cross-dataset}

\begin{table}[!htbp]
\centering
\small
\caption{Cross-dataset composition sidecar. Values are checkpoint-best Top-1, mean $\pm$ sample s.d. The table compares the routing-plus-bias stack, spectral-fusion routing stack, and sampled spectral-fusion variant across the Imagenette and CIFAR-10 lanes; the sampled-minus-spectral-fusion row reports the state-dependent sign of denser fitting.}
\label{tab:cifar-cross-dataset}
\begin{tabular*}{\textwidth}{@{\extracolsep{\fill}}lccc}
\toprule
Composition & Imagenette ($d=256$) & CIFAR-10 ($d=128$) & CIFAR-10 ($d=256$) \\
\midrule
Anchor & $60.65 \pm 0.83$ & $57.44 \pm 1.84$ & $62.16 \pm 0.05$ \\
routing-plus-bias stack & $62.32 \pm 0.27$ & $61.80 \pm 0.06$ & $64.25 \pm 0.24$ \\
Spectral-fusion stack & $62.10 \pm 0.65$ & $62.38 \pm 0.36$ & $63.91 \pm 0.12$ ($n=2$) \\
Sampled spectral-fusion & $61.97 \pm 0.34$ & $64.46 \pm 0.19$ & $67.16 \pm 0.36$ \\
\midrule
$\Delta$ (sampled $-$ spectral-fusion) & $-0.13$ pp (drop) & $\mathbf{+2.08}$ pp (boost) & $\mathbf{+3.25}$ pp (boost) \\
\bottomrule
\end{tabular*}
\end{table}

\FloatBarrier

\subsection{Clean routing-plus-bias stack versus modular null}
\label{app:modular-null-clean-stack}
\begin{figure}[!htbp]
\centering
\includegraphics[width=\textwidth]{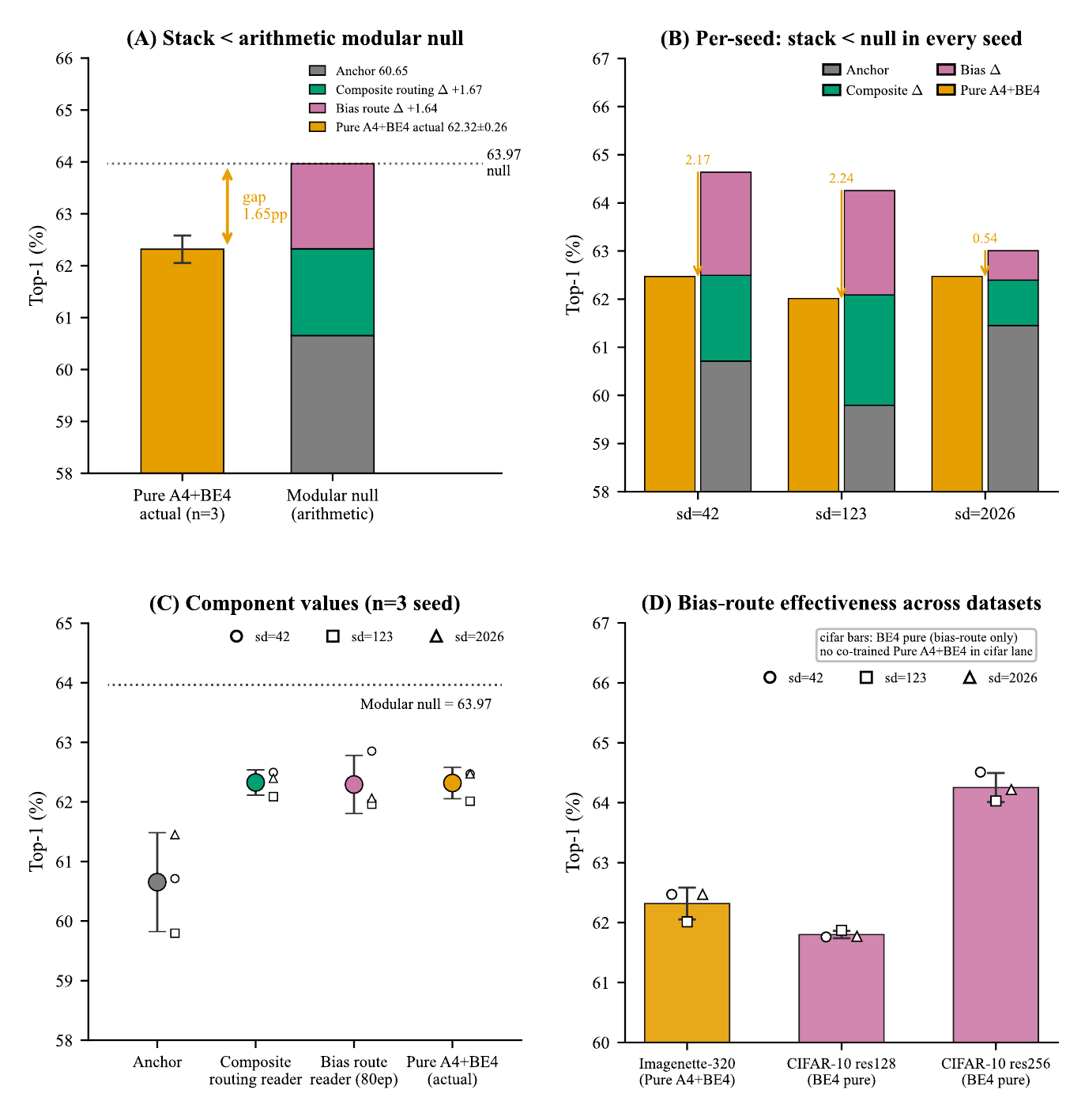}
\caption{Modular-null and clean-stack audit. The clean routing-plus-bias stack remains below the arithmetic modular null ($63.97=60.65+1.68+1.64$), indicating that routing and bias-route gains are not detachable modules.}
\label{fig:app-modularity-audit}
\end{figure}
\FloatBarrier

\subsection{Sampling state-dependence}

\begin{table}[!htbp]
\centering
\small
\caption{Sampling-density state dependence. Denser fitting helps the routing-enhanced state on Imagenette but reverses on the spectral-fusion routing stack. CIFAR-10 lanes are summarized in \tabref{tab:cifar-cross-dataset}.}
\label{tab:sampling-state-dependence}
\begin{tabular*}{\textwidth}{@{\extracolsep{\fill}}p{0.40\textwidth}cc}
\toprule
State & per-seed Top-1 (\%) & mean $\pm$ sample s.d. \\
\midrule
Routing-enhanced reader & 62.50 / 62.09 / 62.39 & $62.33 \pm 0.21$ \\
Stochastic-fit routing reader & 63.16 / 62.73 / 61.81 & $62.57 \pm 0.69$ \\
Spectral-fusion routing stack & 62.57 / 61.35 / 62.37 & $62.10 \pm 0.65$ \\
Sampled spectral-fusion stack & 61.61 / 62.27 / 62.04 & $61.97 \pm 0.34$ \\
\bottomrule
\end{tabular*}
\end{table}

\FloatBarrier

\section{Coordinate-packaging controls}
\label{app:coord-packaging-full}
This appendix supports \secref{sec:controls}. The reader-only control freezes the emitter-side data and trains fresh readers under four input packagings. Residual-only is the weak cell; residual+shift matches full+shift, so subtracting the shared anchor alone is not sufficient in this frozen reader control.

\subsection{Coordinate-packaging table}
\label{app:coord-packaging-raw}

\begin{table}[!htbp]
\centering
\small
\caption{Coordinate-packaging control with formulas and id$_{95}$ summaries. Top-1 is final-5-epoch mean validation accuracy on a frozen CIFAR-10 $d=128$ emitter. The table reports sample/population s.d. for Top-1 and population s.d. for id$_{95}$. Residual-only is the weak cell, while residual+shift matches the full+shift reference.}
\label{tab:coord-packaging-merged}
\begin{tabular*}{\textwidth}{@{\extracolsep{\fill}}llcc}
\toprule
Packaging & Reader input & \shortstack{Top-1 mean\\$\pm$ sample / pop s.d.} & \shortstack{id$_{95}$ mean\\$\pm$ pop s.d.} \\
\midrule
Raw full & $\phi_x$ & 46.17 $\pm$ 0.73 / 0.60 & 10.7 $\pm$ 1.2 \\
Full+shift & $\lambda(\phi_x+\beta)$ & 47.30 $\pm$ 0.32 / 0.26 & 25.7 $\pm$ 0.9 \\
Residual-only & $\lambda(\phi_x-\theta)$ & 43.69 $\pm$ 0.28 / 0.23 & 57.0 $\pm$ 0.0 \\
Residual+shift & $\lambda(\phi_x-\theta+\beta)$ & 47.21 $\pm$ 0.63 / 0.51 & 55.0 $\pm$ 0.8 \\
\bottomrule
\end{tabular*}
\end{table}

\FloatBarrier

No completed $k$-NN neighborhood-size sweep or distance-metric sweep is used to support a main claim. The available boundary reader-convention checks are reported in \appref{app:function-response-full}. This distinction prevents the bias-encoder width sweep in \appref{app:b-encoder-svd} from being mistaken for a nearest-neighbor $k$-sweep.
\FloatBarrier

\clearpage

\end{document}